\pdfoutput=1

\documentclass[11pt]{article}

\usepackage[]{ACL2025}

\usepackage{times}
\usepackage{latexsym}

\usepackage{algorithm}
\usepackage{microtype}
\usepackage{microtype}
\usepackage{times}
\usepackage{latexsym}
\usepackage{amssymb}
\usepackage{amsfonts}
\usepackage{amsmath} 
\usepackage{amsthm} 
\usepackage{booktabs}
\usepackage{enumerate}
\usepackage{enumitem}
\usepackage{graphicx}
\usepackage{subfigure}
\usepackage{xspace}
\usepackage{float}
\usepackage{bbm}
\usepackage{bm}
\usepackage{multirow}
\usepackage{booktabs}
\usepackage{color}
\usepackage{framed}
\usepackage{stfloats}
\usepackage{iitem}
\usepackage{makecell}
\usepackage{float}
\usepackage{placeins}
\usepackage{afterpage}
\usepackage{bbding}
\usepackage[normalem]{ulem}
\usepackage{soul}
\usepackage{xcolor}
\usepackage{color}
\usepackage{colortbl}
\usepackage{algorithm}
\usepackage{algorithmic}
\usepackage{array}
\usepackage{hyperref}
\usepackage{listings}
\usepackage{lipsum,mwe,cuted}
\newcommand{\paratitle}[1]{\vspace{1.5ex}\noindent\textbf{#1}}
\newcommand{\ie}{i.e.,\xspace}

\newcommand{\ignore}[1]{}

\newcommand{\headercolor}{\rowcolor{gray!15}}

\usepackage[T1]{fontenc}
\usepackage{xcolor}
\definecolor{gold}{RGB}{205,133,63}
\definecolor{fGreen}{RGB}{34,139,34}
\definecolor{tOrange}{RGB}{255,165,0}
\definecolor{tBlue}{RGB}{135,206,250}
\definecolor{tPink}{RGB}{255,204,204}
\definecolor{tGreen}{RGB}{205,230,199}
\definecolor{tGold}{RGB}{255,215,0}
\usepackage[utf8]{inputenc}

\usepackage{microtype}

\usepackage{inconsolata}

\title{\texttt{Think\&Cite}: Improving Attributed Text Generation with \\Self-Guided Tree Search and Progress Reward Modeling}

\author{Junyi Li \and Hwee Tou Ng \\
        Department of Computer Science, National University of Singapore \\ \texttt{junyi\_cs@nus.edu.sg, nght@comp.nus.edu.sg}}

\begin{document}
\maketitle

\begin{abstract}

Despite their outstanding capabilities, large language models (LLMs) are prone to hallucination and producing factually incorrect information. This challenge has spurred efforts in attributed text generation, which prompts LLMs to generate content with supporting evidence. In this paper, we propose a novel framework, called \textbf{Think\&Cite}, and formulate attributed text generation as a multi-step reasoning problem integrated with search. Specifically, we propose Self-Guided Monte Carlo Tree Search (SG-MCTS), which capitalizes on the self-reflection capability of LLMs to reason about the intermediate states of MCTS for guiding the tree expansion process. To provide reliable and comprehensive feedback, we introduce Progress Reward Modeling to measure the progress of tree search from the root to the current state from two aspects, i.e., generation and attribution progress. We conduct extensive experiments on three datasets and the results show that our approach significantly outperforms baseline approaches.\footnote{Our dataset and source code are available at \url{https://github.com/nusnlp/Think-Cite}.}

\end{abstract}

\section{Introduction}
\label{introduction}

\begin{figure}[t]
    \centering
    \includegraphics[width=0.48\textwidth]{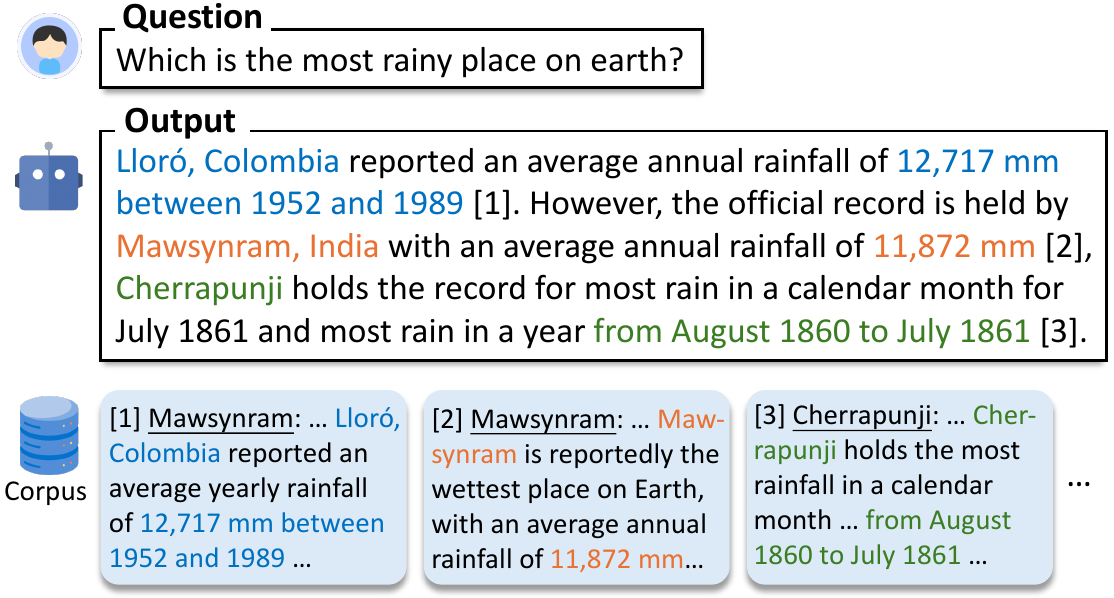}
    \caption{Given a question, the model generates texts by citing passages from a corpus as supporting evidence.} 
    \label{fig:example}
    \vspace{-0.3cm}
\end{figure}

Large language models (LLMs)~\citep{llm_survey} have achieved outstanding performance on many natural language processing tasks. Despite the advances, LLMs often generate responses that contain hallucinations and inaccurate information~\citep{JiLFYSXIBMF23,abs-2311-05232,abs-2309-01219}. This issue undermines their reliability, and more importantly, hurts users' trust in LLMs. To improve the reliability of LLMs, a new paradigm for generation, \emph{attributed text generation}, is proposed, such that LLMs generate responses with in-text citations that provide evidence for any statement~\citep{GaoYYC23}, as shown in Figure~\ref{fig:example}.

Most existing work~\citep{SlobodkinHCSD24,SunCWHWWZY24,FierroAHCMNL24} simply prompts LLMs to provide citations while generating texts. Besides, other work~\citep{LiSHLH0Z24,HuangWHW24} attempts to fine-tune LLMs on massive supervised training data that contains texts with annotated citations. Despite these recent efforts, it remains an open challenge to develop LLMs capable of learning to generate faithful content with reliable references. First, existing approaches adopt an auto-regressive generation paradigm that can be characterized as ``System 1'', a mode of thinking which is fast and instinctive, but less accurate~\citep{kahneman2011thinking}. Thus, any intermediate generation errors (e.g., false statements or erroneous citations) can potentially lead to incorrect final responses. Inspired by research on complex reasoning~\citep{zhang2024llama,abs-2406-14283}, we aim to develop models in the ``System 2'' mode for attribution to external evidence, requiring more in-depth, deliberative, and logical thinking~\citep{kahneman2011thinking}. Second, attributed text generation often involves long text generation. \citet{LiuZL23} find that long-form responses from existing LLMs usually contain unsupported statements and inaccurate citations. We argue that the absence of explicit generation planning in previous work hinders advances in such systems.

In this paper, we propose \textbf{Think\&Cite}, a novel framework integrating search algorithms into attributed text generation. We formulate the task as a multi-step reasoning problem, where the model generates one sentence in each step through an iterative \emph{think-verbalize-cite} paradigm. To enhance this generation process, we propose Self-Guided Monte Carlo Tree Search (SG-MCTS), which extends the classic MCTS with two innovations. 
First, our approach leverages the self-reflection capability of LLMs to deliberate on the \emph{intermediate states of MCTS} in real time, so as to guide the tree expansion process and proactively avoid inadequate reasoning paths. This is different from prior work which mainly reflected on the final outcome or complete trajectory. 
Second, we propose Progress Reward Modeling (PRM) to measure \emph{the progress of tree search} from the root to the current state from two aspects, i.e., generation progress and attribution progress. In contrast to only evaluating single steps, progress-based reward modeling can provide reliable and comprehensive evaluation to guide the MCTS search process.

To the best of our knowledge, we are the first to apply tree search algorithms to the task of attributed text generation. We conduct extensive experiments on three datasets to verify the effectiveness of our approach. 
The results show that our model significantly outperforms previous prompting-based and fine-tuning baselines.

\section{Related Work}

\paratitle{Attributed Text Generation.} Large language models (LLMs) have been used in attributed text generation due to their outstanding language generation capabilities~\citep{GaoYYC23,HuangWHW24,SunCWHWWZY24,LiSHLH0Z24,SlobodkinHCSD24}. The work on LLMs for attributed text generation can be broadly categorized into two types. The first type involves fine-tuning LLMs with preference learning~\citep{LiSHLH0Z24} and reinforcement learning~\citep{HuangWHW24}, which teach LLMs to generate supportive and relevant citations to achieve higher rewards. 
However, this approach depends on human labor to curate high-quality datasets with annotated in-text citations. 
Another line of work directly instructs LLMs to generate attributed texts with appropriate prompts by attribute-then-generate planning~\citep{SlobodkinHCSD24}, or employing external verifiers to guide generation~\citep{SunCWHWWZY24}. However, this approach generates texts and citations in an auto-regressive manner, where any inaccurate intermediate generation can easily lead to failure in the subsequent process. In contrast, our approach proposes self-guided tree search with progressive reward to consider multiple paths.

\paratitle{LLMs with Tree Search.}
Integrating tree search algorithms with LLMs has attracted significant attention. Recent studies have investigated the use of tree search methods to enhance the performance of LLMs during inference~\citep{zhang2024llama,abs-2406-14283,ye-ng-2024-preference}. \citet{sutton2019bitter} highlights the superiority of scaling in both learning and search, over other approaches. Empirical evidence further demonstrates that scaling inference-time computation can significantly improve LLM performance without requiring additional training~\citep{abs-2407-21787,abs-2408-03314}. 
A* search~\citep{HartNR68} and Monte Carlo Tree Search (MCTS)~\citep{BrownePWLCRTPSC12} are employed as planning techniques to improve the performance of LLMs in solving complex reasoning problems. 
{\citet{LewisPPPKGKLYR020} introduces the retrieval score in sequence likelihood to facilitate tree search and Self-RAG~\cite{AsaiWWSH24} explicitly supports tree-decoding with critique tokens.}
Our work is the first to apply tree search algorithms (i.e., Monte Carlo Tree Search) to solve the task of attributed text generation. Moreover, we propose self-guided MCTS that relies on the reflection capability of LLMs to improve tree expansion.

\section{Problem Formulation}
\label{preliminary}

\begin{figure*}[t]
    \centering
    \includegraphics[width=\textwidth]{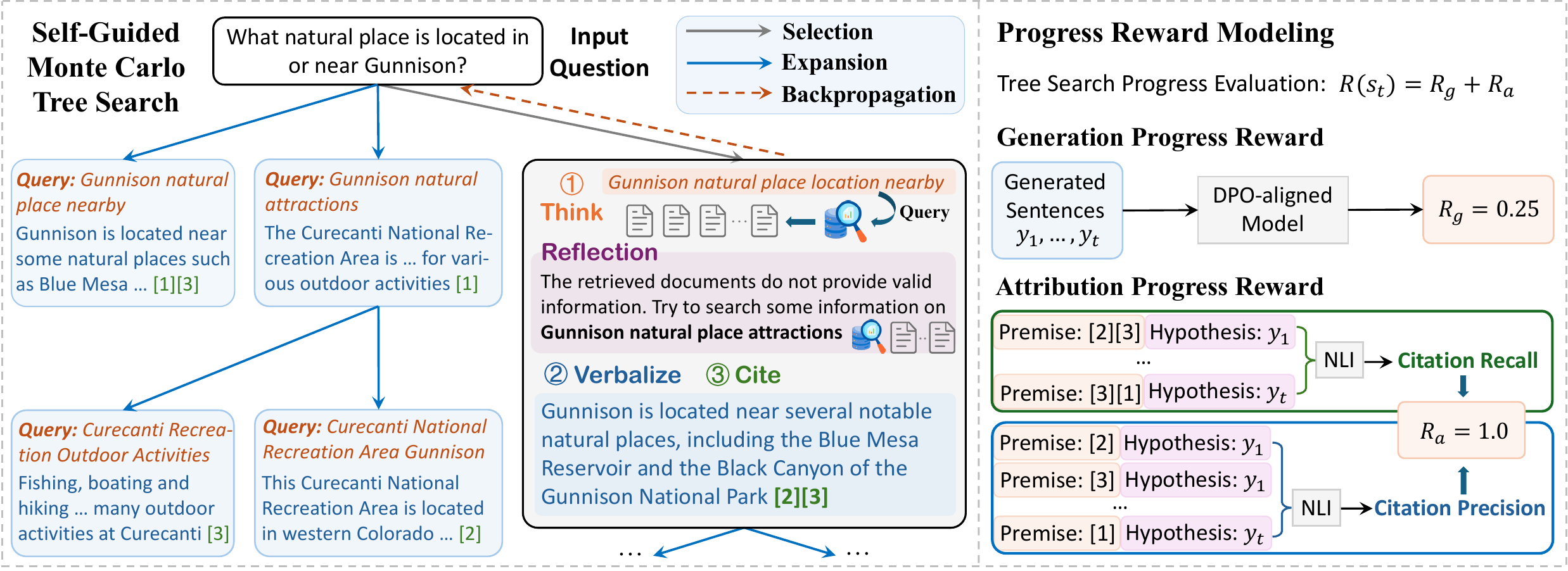}
    \caption{Overall framework of our proposed Think\&Cite approach.
    } 
    \label{fig:model}
    \vspace{-0.3cm}
\end{figure*}

Our proposed framework aims to have a pre-trained LLM $\mathcal{M}_\theta$ generate responses with in-text citations that serve as evidence for the output content, referred to as \emph{attributed text generation}~\citep{SlobodkinHCSD24,GaoDPCCFZLLJG23}. 

Formally, given an input question $\bm{x}$ and a corpus of text passages $\mathcal{D}$, the model $\mathcal{M}_\theta$ is required to generate a response $\bm{y} = \langle y_1,...,y_T \rangle$ consisting of $T$ sentences where each sentence $y_t$ cites a list of passages from $\mathcal{D}$, denoted by $\mathcal{C}_t = \{c_{t,1},...,c_{t,m}\}$. Due to the marginal benefit of incorporating more citations~\citep{GaoYYC23}, in this paper, we allow at most three citations for each sentence ($m \leq 3$), and these citations are enclosed in square brackets, such as \texttt{[1][2]}.
We also mainly focus on knowledge-intensive scenarios where the question concerns world knowledge and most sentences from LLMs contain multiple facts and require supporting citations as evidence. Following prior work~\citep{GaoYYC23,sphere}, we divide the corpus $\mathcal{D}$ into 100-word passages for fine-grained retrieval, which makes it easier for humans to verify and does not introduce too much irrelevant information.

\section{Approach}
\label{approach}

The proposed \textsc{Think\&Cite} framework builds on a language agent for attributed text generation, using Self-Guided Monte Carlo Tree Search (SG-MCTS) to plan and search over multiple generation paths, and Progress Reward Modeling to provide progressive signals for the search process. Figure~\ref{fig:model} depicts the overall framework of our approach.

\subsection{Attributed Text Generation Agent}
\label{sec-agent}

Inspired by prior work~\citep{Yao-arxiv-2022-ReAct,fireact}, we develop a language agent to address the task of attributed text generation, which performs an iterative \emph{think-verbalize-cite} process, leveraging the reasoning and planning capabilities of LLMs.

\paratitle{Iterative Think-Verbalize-Cite.} 
{In our approach, the language agent generates one sentence in each step.}
To generate the $t$-th sentence, the agent first proactively thinks about the blueprint, such as topic or abstract~\cite{NarayanMAGLH00L23}, for the next sentence that serves as a retrieval query $q_t$. Then the agent uses the search tool to retrieve the most relevant top-$K$ passages $\mathcal{D}_t$ from the given corpus $\mathcal{D}$ through the \texttt{Search} action, \ie ``\texttt{Search: \{query\}}''. Based on the retrieved passages, the agent verbalizes a sentence $y_t$ by citing a list of passages $\mathcal{C}_t$ from $\mathcal{D}_t$ through the \texttt{Generate} action, \ie ``\texttt{Generate: \{sentence\}}''. The historical queries, retrieved passages, generated sentences, and citations, denoted as $\mathcal{H} = \{\langle q_i, \mathcal{D}_i, y_i, \mathcal{C}_i \rangle \}_{i=1}^t$ will be combined as the context for thinking and verbalizing in the next step. If the agent believes that the task has been solved, it can output ``\texttt{End}'' to terminate this process. 
In this way, the agent deliberately plans and retrieves diverse information, which can dynamically consider shift in content focus as the generation progresses, in contrast to prior work relying on a static reference corpus~\citep{SlobodkinHCSD24,HuangWHW24,LiSHLH0Z24,FierroAHCMNL24}. 
Besides, this paradigm is similar to recent work on iterative retrieval-augmented generation~\citep{JiangXGSLDYCN23,ShaoGSHDC23}, but differs in that our work requires the model to anticipate a content blueprint for the next generation to retrieve relevant information and carefully select appropriate references for incorporating them at suitable positions within the generated text.

\subsection{Self-Guided Monte Carlo Tree Search} 
\label{sec-reflection}

We formulate attributed text generation as a multi-step reasoning problem where the model deliberates on the attribution for text.
Monte Carlo Tree Search has become an effective search algorithm for many decision-making tasks~\citep{alphago,YeLKAG21}.
In this work, we propose \emph{Self-Guided Monte Carlo Tree Search (SG-MCTS)}, which capitalizes on the self-reflection capability of LLMs to guide the search process of MCTS.
Previous work~\citep{ShinnCGNY23,ZhouYSWW24,yu2024improving} often reflects on the final outcome or the complete trajectory, which is inefficient and sparse. In contrast, our method aims to criticize and deliberate on \emph{intermediate states} of MCTS to guide tree expansion in real time and proactively ignore erroneous generation paths.

Generally, MCTS builds a search tree $\mathcal{T}$ based on a policy model $\pi_\theta$, which is often the LLM $\mathcal{M}_\theta$. In this tree, node $s_t = [q_t, \mathcal{D}_t, y_t, \mathcal{C}_t]$ denotes a state at the $t$-th tree level, including search query $q_t$, retrieved passages $\mathcal{D}_t$, generated sentence $y_t$, and cited passages $\mathcal{C}_t$. 
The root node $s_0 = [\bm{x}]$ denotes the input question. 
{Each node contains one sentence and the final output is the concatenation of sentences $\langle y_1,...,y_T \rangle$, where each sentence $y_t$ comes from a node on the path from the root node to the leaf node.}
In each iteration, SG-MCTS follows four steps, i.e., selection, reflection-guided expansion, evaluation, and backpropagation. 

\paratitle{Selection Phase.} The selection stage aims to identify a node $s_t$ from the search tree $\mathcal{T}$ for subsequent expansion. The Upper Confidence Bound applied to Trees (UCT) algorithm~\citep{uct} is employed to select the optimal node with the highest UCT score:
\begin{equation}
    \label{eq:uct}
    {UCT}(s_t) = V(s_t) + w\sqrt{\frac{\ln{N(p)}}{N(s_t)}},
\end{equation}
where $V(s_t)$ is the value function (expected reward) of $s_t$ estimated at the evaluation stage, $N(s_t)$ is the visit count of $s_t$, $w$ is a weight controlling exploration, and $p$ is the parent node of $s_t$. 

\paratitle{Reflection-Guided Expansion Phase.} In the expansion phase, the selected node $s_t$ is expanded by generating successor node $s_{t+1}$ through the think-verbalize-cite process. The \textbf{Think} step first generates an initial search query $\hat{q}_{t+1}$, abstracting the topic or content of the next sentence, which will be used to retrieve passages $\widehat{\mathcal{D}}_{t+1}$.
{However, the initial query might be obscure or inaccurate, which can hinder subsequent evidence retrieval and ultimately result in incorrect sentence generation. Moreover, some questions do not have straightforward answers, necessitating iterative refinement to retrieve and generate accurate results. }
Therefore, we introduce the \textbf{Reflection} step, where the model reflects on the initial query $\hat{q}_{t+1}$ to identify errors based on the question $\bm{x}$ and retrieved passages $\widehat{\mathcal{D}}_{t+1}$ as:
\begin{equation}
    u = \mathcal{M}_\theta(\hat{q}_{t+1}, \widehat{\mathcal{D}}_{t+1}, \bm{x}),
\end{equation}
where the reflection text $u$ includes retrieval advice on certain aspects, e.g., the query should be more focused in search topics. Based on the reflection, the policy model reformulates a new query $q_{t+1}$ to retrieve more relevant passages $\mathcal{D}_{t+1}$:
\begin{equation}
    q_{t+1}, \mathcal{D}_{t+1} = \mathcal{M}_\theta(u, \hat{q}_{t+1}, \widehat{\mathcal{D}}_{t+1}).
\end{equation}
{Note that the reflection step can be iterated until the model determines that the retrieved evidence is supportive or the maximum number of reflection steps is reached.} Finally, the \textbf{Verbalize} and \textbf{Cite} steps generate the next sentence $y_{t+1}$ with accurate citations $\mathcal{C}_{t+1}$ from $\mathcal{D}_{t+1}$ as:
\begin{equation}
    y_{t+1}, \mathcal{C}_{t+1} = \mathcal{M}_\theta (q_{t+1}, \mathcal{D}_{t+1}, \mathcal{H}),
\end{equation}
where $\mathcal{H}$ is the historical context. The new node includes search query, retrieved corpus, generated sentence, and cited passages, denoted as $s_{t+1} = [q_{t+1}, \mathcal{D}_{t+1}, y_{t+1}, \mathcal{C}_{t+1}]$. Compared to simple expansion in typical MCTS, our approach can refine flawed expanded nodes to avoid low-quality generation. Since the MCTS tree is built step by step, improving the quality of the next action allows the model to navigate more favorable pathways in the vast search space, thereby enhancing the overall search quality and efficiency of the tree. 

\paratitle{Evaluation Phase.} The evaluation stage aims to compute the expected reward $R(s_{t+1})$ of the newly expanded node $s_{t+1}$ using Progress Reward Modeling (see Section~\ref{sec-reward}). The progress evaluation involves two aspects: generation and attribution. The generation progress reward $R_g$ measures the \emph{textual quality} of generated sentences so far $y_1,...,y_{t+1}$. The attribution progress reward $R_a$ evaluates the \emph{attribution consistency} between generated sentences $y_1,...,y_{t+1}$ and cited passages $\mathcal{C}_1,...,\mathcal{C}_{t+1}$. Finally, the total reward is computed as the sum of both: $R(s_{t+1}) = R_g + R_a$.

\paratitle{Backpropagation Phase.} In the backpropagation phase, the reward $R(s_{t+1})$ of the new node is propagated back to its parent node $s_t$, updating the value function of each node $s_0, s_1, ..., s_t$ along the path from the root node to its parent node:
\begin{align}
    \label{eq:back}
    N_\text{new}(s_i) &= N_\text{old}(s_i)+1,~~0 \leq i \leq t \\
    V_\text{new}(s_i) &= \frac{V_\text{old}(s_i)N_\text{old}(s_i)+R(s_{t+1})}{N_\text{new}(s_i)},
\end{align}
where $N_\text{old}(s_i)$ and $V_\text{old}(s_i)$ are the prior visit count and value function of node $s_i$, respectively.

The above four steps are performed iteratively until the policy model outputs ``\texttt{End}'', indicating the task has been solved or the maximum number of MCTS iterations is reached.

\subsection{Progress Reward Modeling}
\label{sec-reward}

Previous outcome reward models~\citep{abs-2402-06457} and process reward models \citep{LightmanKBEBLLS24} mainly evaluate the final result or intermediate steps. In this work, we propose to measure the progress of tree search from the root $s_0$ to state $s_{t+1}$ after taking the next step. Since an attributed text includes the text and its citations, we design two aspects for progress reward modeling, \emph{Generation Progress Reward} and \emph{Attribution Progress Reward}, to evaluate the quality of the generated textual content and the relevance of citations, respectively.

\subsubsection{Generation Progress Reward} 

In direct preference optimization (DPO)~\citep{dpo}, the token-level log-ratio can be explained as an implicit token-level reward under a max-entropy reinforcement learning (RL) formulation. Thus, we propose to leverage existing DPO-aligned models to 
measure the quality score $R_g$ of the generated sentences $\bm{y}_{1:t+1} = y_1,...,y_{t+1}$ after generating the next sentence $y_{t+1}$.

Specifically, we define a sentence-level Markov Decision Process (MDP) where the state $\bm{s}_t = \langle \bm{x}, y_1,...,y_t \rangle$ denotes the input and sentences generated so far and the initial state $\bm{s}_0 = \bm{x}$ is the input question. The action $\bm{a}_t = y_{t+1}$ denotes the next sentence to be generated. Hence, the RLHF optimization objective can be rewritten as a max-entropy RL problem at the sentence level:
\begin{equation}
    \mathbb{E}_{\bm{a}_t \sim \pi_\theta(\cdot|\bm{s}_t)} [\sum_{t=1}^T r'(\bm{s}_t,\bm{a}_t)] + \beta\mathbb{E}_{\bm{s}_0 \sim \mathcal{X}}[\mathcal{H}(\pi_\theta(\cdot|\bm{s}_0))], \nonumber
\end{equation}
where the sentence-level reward function $r'$ can be calculated as:
\begin{align}
    &r'(\bm{s}_t, \bm{a}_t) = \nonumber \\ 
    &\left\{ 
    \begin{aligned}
        &\beta \log \pi_\text{ref}(\bm{a}_t|\bm{s}_t), ~\text{if}~\bm{s}_{t+1}~\text{is not terminal,}\\
        & r(\bm{y}|\bm{x})+ \beta\log \pi_\text{ref}(\bm{a}_t|\bm{s}_t)~\text{if}~\bm{s}_{t+1}~\text{is terminal.}
    \end{aligned}
    \right. \notag
\end{align}
The max-entropy RL formulation derives the optimal value function $V^*$ and $Q$-function $Q^*$ as:
\begin{align}
    Q^*(\bm{s}_t,\bm{a}_t) &= r'(\bm{s}_t,\bm{a}_t) + V^*(\bm{s}_{t+1}), \notag \\
    V^*(\bm{s}_t) &= \log \sum_{\bm{a}} \exp(Q^*(\bm{s}_t,\bm{a})), \text{when}~t \leq T. \notag
\end{align}
Thus, the optimal policy $\pi^*$ is derived as:
\begin{align}
    \beta \log \pi^*(\bm{a}_t|\bm{s}_t) &= Q^*(\bm{s}_t,\bm{a}_t) - V^*(\bm{s}_t), \notag \\
    \Rightarrow \quad \beta \log \frac{\pi^*(\bm{a}_t|\bm{s}_t)}{\pi_\text{ref}(\bm{a}_t|\bm{s}_t)} &= V^*(\bm{s}_{t+1}) - V^*(\bm{s}_t). \notag
\end{align}
This motivates us to use a DPO policy to derive the partial sum of the reward to formulate the progress reward $R_g$ for a partial response $\bm{y}_{1:t+1}$:
\begin{align}
    \sum_{k=0}^{t} \beta \log \frac{\pi^*(\bm{a}_k|\bm{s}_k)}{\pi_\text{ref}(\bm{a}_k|\bm{s}_k)} = V^*(\bm{s}_{t+1}) - V^*(\bm{s}_0), \notag \\
    \Rightarrow \quad R_g(\bm{y}_{1:t+1}) = \sum_{k=0}^{t} w_k \log \frac{\pi^*(y_{k+1}|\bm{x},\bm{y}_{1:k})}{\pi_\text{ref}(y_{k+1}|\bm{x},\bm{y}_{1:k})}, \notag
\end{align}
where $\bm{y}_{1:k}=y_1,...,y_k$, $w_k = \frac{1}{|\bm{y}_{1:k}|}$ is the weight for each sentence-level log-likelihood ratio, and {$|\bm{y}_{1:k}|$ is the number of tokens in $\bm{y}_{1:k}$}.

\subsubsection{Attribution Progress Reward} 

We employ two citation metrics used in prior work~\citep{GaoYYC23}, i.e., citation recall and precision, for attribution progress reward $R_a$. 

Specifically, citation recall measures the percentage of sentences in the partial response $\bm{y}_{1:t+1}$ that can be supported by the corresponding cited passages. We employ an NLI model~\citep{HonovichAHTKCSS22} to examine whether the cited passages can entail the model response. For each sentence $y_i$ ($1 \leq i \leq t+1$), we concatenate the cited passages in $\mathcal{C}_i$ as premise and regard the generated sentence $y_i$ as hypothesis for the NLI model. If the premise entails the hypothesis, we set the citation recall as 1, and 0 otherwise. Citation precision evaluates the percentage of citations that support the corresponding sentence. We use the same NLI model above to calculate the precision score. 
For each citation $c_{i,j}$, its precision score is set to 1 if (1) all citations in $\mathcal{C}_i$ entail the generated sentence $y_i$ and (2) $\mathcal{C}_i \setminus \{ c_{i,j} \}$ does not entail the sentence $y_i$. Otherwise, the precision score is set to 0. We compute the precision score for each citation (0 or 1) and average over all citations. Finally, we calculate F1 score as the attribution progress reward $R_a(\bm{y}_{1:t+1}, \mathcal{C}_1,...,\mathcal{C}_{t+1})$ to provide a balanced attribution quality measure between the generated sentences and cited passages.

\section{Experiments}
\label{experiment}

\subsection{Experimental Setup}
\label{sec:app_experiment}

\paratitle{Datasets.} For evaluation, we use the ALCE benchmark~\citep{GaoYYC23}, which consists of three datasets: (1) \textbf{ASQA}~\citep{StelmakhLDC22}, a long-form QA dataset containing ambiguous questions that require multiple  answers to cover different aspects; (2) \textbf{QAMPARI}~\citep{qampari}, a factoid QA dataset where the answer to each question is a list of entities drawn from different passages; (3) \textbf{ELI5}~\citep{FanJPGWA19}, a long-form QA dataset containing how/why/what questions. 
For ASQA and QAMPARI, most questions can be answered by Wikipedia, thus we adopt the 2018/12/20 Wikipedia snapshot as the corpus. For ELI5, since its questions are diverse in topics, we use Sphere~\citep{sphere}, a filtered version of Common Crawl, as the corpus. Following~\citet{GaoYYC23}, we adopt GTR~\citep{gtr} for Wikipedia and BM25~\citep{bm25} for Sphere to retrieve the top $100$ passages as the corpus for each question. See Appendix~\ref{sec:app-dataset} for more details.

\paratitle{Evaluation Metrics.} We use the evaluation metrics in the original ALCE benchmark. 
To evaluate the correctness of the output, we use \textbf{Exact Match (EM) Recall} for ASQA, \textbf{Recall-5} for QAMPARI, and \textbf{Claim Recall} for ELI5, for measuring the percentage of gold answers (key information pieces) in the output. We further compute \textbf{Precision} as a correctness metric for the QAMPARI dataset, measuring the percentage of generated answers that are correct. To evaluate the citation quality of the output, we compute \textbf{Citation Recall}, which measures the percentage of sentences in the output that can be entailed from their cited passages, and \textbf{Citation Precision}, which measures the percentage of citations that can help support the output sentences.

\paratitle{Baselines.} We compare our approach to the following baselines based on ChatGPT and GPT-4o:

\textbullet~\textbf{Vanilla RAG} directly instructs the model to generate responses and cite accordingly based on the given top $5$ passages. We use in-context learning~\citep{gpt3} with two demonstrations.

\textbullet~\textbf{Summary/Snippet RAG} provides summaries or snippets of passages instead of the full text. The model will generate responses with citations based on the top $10$ passage summaries or snippets. 

\textbullet~\textbf{Interact} allows the model to further access the full text of certain passages for the Summary/Snippet RAG method. The model can propose an action ``\texttt{Check: Document [1][2]}'' to obtain the full text of the corresponding documents.

\textbullet~\textbf{Inline Search} allows the model to request an action ``\texttt{Search: \{query\}}'' to retrieve the most relevant passage from the top $100$ passages. This method is similar to our approach which serves as a direct comparison.

\textbullet~\textbf{ReRank} randomly samples four responses for each question and select the best one based on the citation recall metric.

The above baselines have been employed and evaluated on the original ALCE benchmark, as reported in~\citep{GaoYYC23}. Besides, we compare our approach to existing work on attributed text generation. \textbf{FG-Reward}~\citep{HuangWHW24} proposes to use fine-grained rewards as training signals to fine-tune LLaMA-2-7B~\citep{llama2} to generate attributed responses. \textbf{VTG}~\citep{SunCWHWWZY24} guides the LLM (i.e., \texttt{text-davinci-003}) using an evolving memory and a two-tier verifier. \textbf{APO}~\citep{LiSHLH0Z24} curates a preference dataset and uses preference optimization to fine-tune LLaMA-2-13B. Note that our model directly performs inference on the test sets of the three evaluation datasets without performing any fine-tuning.

\paratitle{Implementation Details.} We use LLaMA-3.1-8B-Instruct and GPT-4o as our policy models to assess the performance of our approach. For the reward models, we adopt a DPO model, i.e., Llama-3-8B-SFR-Iterative-DPO-R\footnote{https://huggingface.co/Salesforce/LLaMA-3-8B-SFR-Iterative-DPO-R}, to compute the generation progress reward and an NLI model, i.e., T5-XXL-TRUE-NLI-Mixture~\citep{HonovichAHTKCSS22}, to compute the attribution progress reward. For each search query, we retrieve top $3$ passages from the corpus as the candidate references $\mathcal{D}_t$. In UCT algorithm (Eq.~\ref{eq:uct}), the weight $w$ is set to $0.2$. For SG-MCTS, we expand three child nodes for each parent node and {set the maximum number of reflection steps to $10$}, the maximum tree layer to $6$, and the maximum iteration of MCTS to $30$. 

\begin{table*}[t]
\centering
\small
\begin{tabular}{lcccccccccc}
\toprule
\multirow{3.5}{*}{\textbf{}} & \multicolumn{3}{c}{\textbf{ASQA}} & \multicolumn{4}{c}{\textbf{QAMPARI}} & \multicolumn{3}{c}{\textbf{ELI5}} \\
& \textbf{Correctness} & \multicolumn{2}{c}{\textbf{Citation}} & \multicolumn{2}{c}{\textbf{Correctness}} & \multicolumn{2}{c}{\textbf{Citation}} & \textbf{Correctness} & \multicolumn{2}{c}{\textbf{Citation}} \\
\cmidrule(r){2-4}\cmidrule(r){5-8}\cmidrule(r){9-11}
& \textbf{EM Rec.} & \textbf{Rec.} & \textbf{Prec.} & \textbf{Recall-5} & \textbf{Prec.} & \textbf{Rec.} & \textbf{Prec.} & \textbf{Claim Rec.} & \textbf{Rec.} & \textbf{Prec.} \\
\midrule
\headercolor
\multicolumn{11}{c}{\textbf{ChatGPT}} \\
\specialrule{0em}{1pt}{1pt}
Vanilla RAG & 40.4 & 73.6 & 72.5 & 20.8 & 20.8 & 20.5 & 20.9 & 12.0 & 51.1 & 50.0 \\
~~ w/ ReRank & 40.2 & 84.8 & 81.6 & 22.8 & 21.4 & 21.2 & 21.4 & 11.4 & 69.3 & 67.8 \\
Summary RAG & 43.3 & 68.9 & 61.8 & 23.6 & 21.2 & 23.6 & 25.7 & 12.5 & 51.5 & 48.2 \\
~~ w/ Interact & 39.1 & 73.4 & 66.5 & 23.2 & 20.9 & 22.1 & 24.3 & 13.7 & 50.1 & 49.2 \\
Snippet RAG & 41.4 & 65.3 & 57.4 & 24.5 & 21.5 & 22.9 & 24.9 & 14.3 & 50.4 & 45.1 \\
~~ w/ Interact & 41.2 & 64.5 & 57.7 & 22.4 & 20.8 & 21.6 & 23.1 & 13.3 & 47.8 & 45.2 \\
Inline Search & 32.4 & 58.3 & 58.2 & 17.2 & 20.4 & 14.9 & 14.9 & 13.4 & 45.6 & 43.7 \\
\midrule
\headercolor
\multicolumn{11}{c}{\textbf{GPT-4o}} \\
\specialrule{0em}{1pt}{1pt}
Vanilla RAG & 41.3 & 68.5 & 75.6 & 22.2 & 25.0 & 25.9 & 27.0 & \underline{20.3} & 53.1 & 55.2 \\
~~ w/ ReRank & 42.1 & 83.4 & \underline{82.3} & 32.6 & 30.9 & 33.1 & 32.8 & 19.4 & 68.2 & 65.7 \\
Summary RAG & 46.5 & 70.2 & 67.2 & 36.2 & 34.0 & 36.2 & 39.8 & 18.7 & 64.4 & 63.9 \\
~~ w/ Interact & \underline{48.1} & 73.1 & 72.8 & \underline{38.2} & \underline{37.1} & 39.2 & 40.6 & 20.1 & 69.2 & 66.2 \\
Snippet RAG & 45.1 & 68.9 & 66.5 & 37.1 & 35.2 & 37.2 & 38.3 & 19.8 & 64.8 & 60.1 \\
~~ w/ Interact & 45.2 & 67.8 & 66.7 & 37.2 & 34.5 & 38.7 & 39.5 & 19.9 & 68.1 & 65.1 \\
Inline Search & 40.3 & 65.7 & 66.9 & 27.8 & 27.2 & 19.4 & 23.8 & 12.5 & 50.2 & 53.3 \\
\midrule
FG-Reward & 40.1 & 77.8 & 76.3 & 16.7 & 19.5 & 19.5 & 20.1 & 11.5 & 60.9 & 60.2 \\
VTG & 41.5 & \underline{86.7} & 80.0 & 20.3 & 22.4 & \underline{43.5} & \underline{46.9} & 16.7 & \underline{82.6} & \underline{71.6} \\
APO & 40.5 & 72.8 & 69.6 & 15.4 & 20.6 & 17.5 & 19.3 & 13.5 & 26.0 & 24.5 \\
\midrule 
Ours~(LLaMA) & 45.2 & 82.3 & 80.6 & 25.7 & 28.1 & 40.5 & 43.1 & 17.4 & 77.5 & 75.3 \\
Ours~(GPT-4o) & \textbf{50.1} & \textbf{89.5} & \textbf{87.1} & \textbf{45.2} & \textbf{41.9} & \textbf{50.6} & \textbf{52.8} & \textbf{25.9} & \textbf{85.6} & \textbf{80.2} \\
\bottomrule
\end{tabular}
\caption{Evaluation results on three datasets on attributed text generation. ``Rec.'' and ``Prec.'' are short for recall and precision. {The \textbf{bold} and \underline{underline} fonts denote the best and second best results in each dataset, respectively.}}
\vspace{-0.3cm}
\label{tab:main-result}
\end{table*}

\subsection{Main Results}

Table~\ref{tab:main-result} shows the results of our method and baselines across three datasets.

Firstly, it can be observed that the three retrieval-augmented generation (RAG) methods {fall short compared to more recent models}, while using summaries or snippets improves correctness. This improvement comes at the expense of citation quality as the passage information is highly compressed. ReRank leads to consistent improvements in citation quality across three datasets.
As a direct comparison, Inline Search exhibits worse performance compared to other prompting-based baselines. This is due to simply proposing search queries without considering evidence quality. 

Secondly, by fine-tuning an LLM on supervised training data with annotated citations, FG-Reward and APO show increased citation quality in both ASQA and ELI5 datasets. Besides, VTG employs a generation verifier and a memory verifier to assess the logical support of evidence, leading to strong citation quality (e.g., 86.7\% citation recall in ASQA). However, fine-tuning the LLMs is constrained by the quality and quantity of the supervised training data where the supporting evidence requires substantial costs to link to correct sources. Moreover, these approaches still rely on auto-regressive generation, where any intermediate generation errors (e.g., false statements or inadequate citations) may result in problematic final responses.

Finally, our approach outperforms all baselines significantly across all three datasets. Think\&Cite formulates attributed text generation as a multi-step reasoning problem and introduces a slow and deliberative thinking mode to search for the optimal solutions. Think\&Cite leverages the self-reflection capability of LLMs to guide the tree expansion process. Besides, the proposed progress reward modeling can further provide comprehensive and reliable feedback to help the model explore better generated responses.

\subsection{Further Analysis}

We report further analysis of our method using GPT-4o on ASQA, as we have similar findings in the other datasets.

\begin{table}[t]
\centering
\small
\begin{tabular}{l  c c  c }
    \toprule
    \multirow{2.5}{*}{\textbf{Method}} & \multicolumn{1}{c}{\textbf{Correctness}} & \multicolumn{2}{c}{\textbf{Citation}} \\
    \cmidrule(r){2-4}
    & \textbf{EM Rec.} & \textbf{Rec.} & \textbf{Prec.} \\
    \midrule
    Think\&Cite & 50.1 & 89.5 & 87.1 \\
    \midrule
    w/o SG-MCTS  & 42.1 & 78.2 & 75.0 \\
    w/o Reflection & 46.5 & 83.6 & 80.3 \\
    w/o GP Reward & 47.1 & 86.2 & 84.9 \\
    w/o AP Reward & 46.7 & 81.3 & 80.4 \\
    \bottomrule
\end{tabular}
\caption{Ablation study in ASQA.}
\vspace{-0.5cm}
\label{tab:ablation}
\end{table}

\paratitle{Ablation Study.} To validate the effectiveness of our proposed framework, we conduct an ablation analysis of its key design elements. We design four variants: (1) \emph{w/o SG-MCTS} removes self-guided MCTS and directly generates answers step by step; (2) \emph{w/o Reflection} removes the reflection step and adopts the vanilla MCTS algorithm; (3) \emph{w/o GP Reward} removes the generation progress reward $R_g$; and (4) \emph{w/o AP Reward} removes the attribution progress reward $R_a$.
We show the results in Table~\ref{tab:ablation}. All the variants perform worse than the original method, indicating the effectiveness of each component. Specifically, the performance of \emph{w/o SG-MCTS} drops significantly, indicating that integrating search algorithms in attributed text generation is highly beneficial. Using vanilla MCTS (\emph{w/o Reflection}) results in worse citation quality, due to the introduction of erroneous references without reflection on the retrieved results. Similarly, both \emph{w/o GP Reward} and \emph{w/o AP Reward} lead to worse performance, indicating that both generation and citation quality check are critical.

\begin{figure}[t]
    \centering
    \small
    \includegraphics[width=0.48\textwidth]{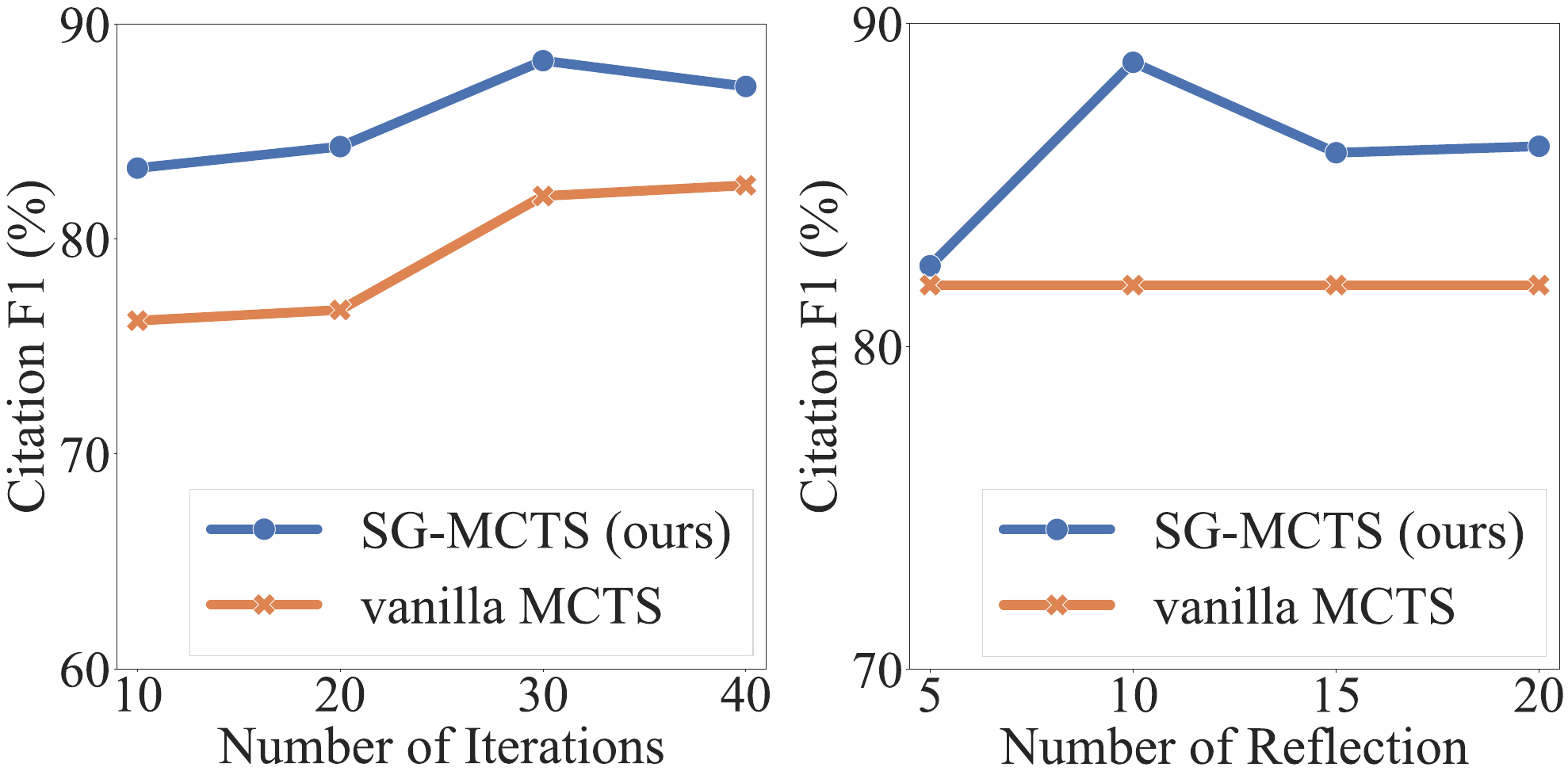}
    \caption{Results on ASQA \emph{w.r.t.} the number of iterations (\textbf{left}) or the number of reflection steps (\textbf{right}).
    } 
    \label{fig:simulation_acc}
    \vspace{-0.3cm}
\end{figure}

\paratitle{Reflection vs {MCTS Iteration}.}
In each iteration, SG-MCTS employs four key steps and reflection to improve the quality of intermediate states in the expansion phase by criticizing and refining error queries. To examine the effectiveness of reflection, we compare the performance between increasing the maximum number of MCTS iterations and reflection steps. We first vary the maximum number of MCTS iterations in $\{10, 20, 30, 40\}$ and fix the maximum number of reflection steps as $10$. Similarly, we also vary the maximum number of reflection steps in $\{5, 10, 15, 20\}$ and fix the maximum number of iterations as $30$. We present the F1 score based on the citation recall and precision in Figure~\ref{fig:simulation_acc}. The figure shows that increasing the number of iterations and reflection steps enhance the task performance. This is expected, as more extensive exploration raises the probability of finding the correct generation. However, more reflection steps make the model ``overthinks'', introducing noise and resulting in performance degradation. SG-MCTS outperforms vanilla MCTS without reflection, since incorrect retrieval is likely to exist in parent nodes, causing the reasoning process of child nodes to continue along the wrong path. The reflection step refines flawed retrieval resulting from inadequate queries, allowing subsequent exploration to proceed more accurately.

\begin{figure}[t]
    \centering
    \small
    \includegraphics[width=0.48\textwidth]{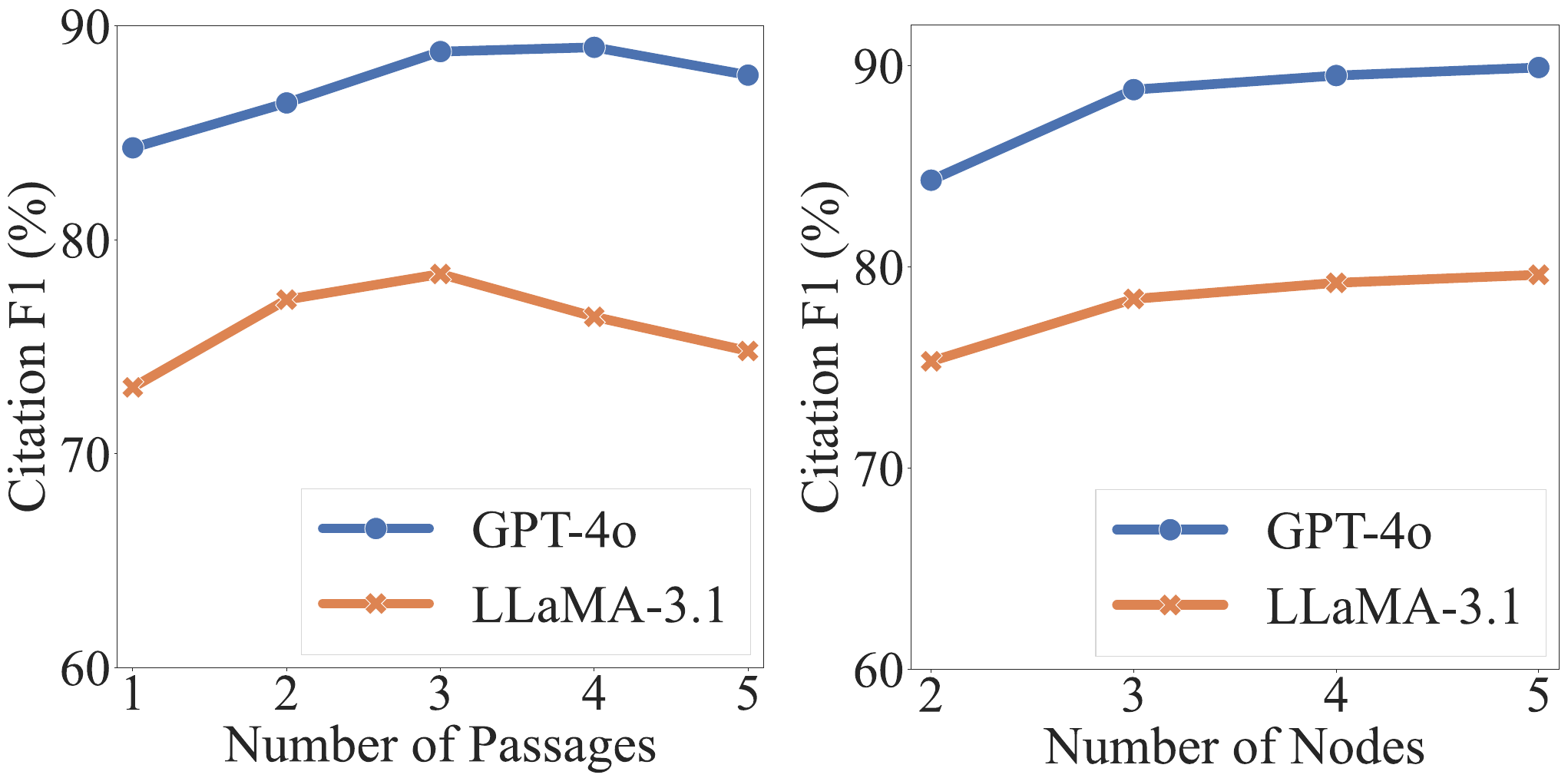}
    \caption{Results on ASQA \emph{w.r.t.} the number of passages (\textbf{left}) or the number of expanded nodes (\textbf{right}).
    } 
    \label{fig:retrieval_expansion}
    \vspace{-0.3cm}
\end{figure}

\paratitle{Hyper-parameter Analysis.} There are two critical hyper-parameters for correctness and citation quality: the number of retrieved passages $\mathcal{D}_t$ for each query $q_t$, and the number of expanded child nodes $s_{t+1}$ in tree search. As illustrated in Figure~\ref{fig:retrieval_expansion}, citation quality can be improved initially with increasing number of retrieved passages. However, further increase beyond a certain threshold results in worse performance, mainly because incorporating more passages introduces noise, negatively impacting the credibility of the generated content. Besides, we observe a consistent improvement when increasing the number of expanded nodes, although the improvement stabilizes later. Since expanding more nodes leads to higher computational costs, we sample three child nodes per parent node.

\subsection{Case Study}

To facilitate understanding of the entire workflow of our approach, we conduct a qualitative analysis in ASQA. We present an example in Figure~\ref{fig:case} in Appendix~\ref{sec:app-case}. Throughout the search process, the LLM considers the input question as the root node and incrementally expands the search tree to reach the terminal state. As shown in the example, the model first generates the query (i.e., ``\emph{Gunnison natural place location nearby}'') to retrieve passages. Since the passages do not contain valid information to answer the question, the model reflects and proposes a new query (i.e., ``\emph{Gunnison natural place attractions}'') for retrieval. Based on the retrieved passages, the model generates the sentence and cites the second and third passages (i.e., ``\texttt{[2][3]}''). By following the multi-step generation process, the model can deliberate on the topic and output reliable content with accurate citations.

\section{Conclusion}
\label{conclusion}

In this work, we proposed Think\&Cite, a novel framework integrating tree search for attributed text generation. Think\&Cite built upon an iterative think-verbalize-cite generation paradigm. To enhance the generation process, we proposed self-guided Monte Carlo Tree Search, which leveraged the self-reflection capability of LLMs to criticize and refine the intermediate states of MCTS to guide tree expansion. Moreover, we proposed progress reward modeling to measure the progress of tree search and to provide reliable feedback. Extensive experiments on three datasets showed that our proposed Think\&Cite outperforms traditional prompting and fine-tuning methods.

\section*{Limitations} \label{limitations}

The scope of our experiments is constrained by the substantial computational cost of tree-based search methods. Future work can explore a broader range of attributed text generation datasets. In our model, Monte Carlo tree search is employed for self-guided generation. Future work can explore additional search algorithms to evaluate the generality and robustness of our proposed framework.

\section*{Acknowledgments}

This work is fully supported by the Advanced Research and Technology Innovation Centre (ARTIC), the National University of Singapore under Grant (project number: ELDT-RP1).

\bibliography{custom,acl2025}

\begin{thebibliography}{46}
\expandafter\ifx\csname natexlab\endcsname\relax\def\natexlab#1{#1}\fi

\bibitem[{Amouyal et~al.(2022)Amouyal, Rubin, Yoran, Wolfson, Herzig, and Berant}]{qampari}
Samuel~Joseph Amouyal, Ohad Rubin, Ori Yoran, Tomer Wolfson, Jonathan Herzig, and Jonathan Berant. 2022.
\newblock {QAMPARI}: An open-domain question answering benchmark for questions with many answers from multiple paragraphs.
\newblock \emph{CoRR}, abs/2205.12665.

\bibitem[{Asai et~al.(2024)Asai, Wu, Wang, Sil, and Hajishirzi}]{AsaiWWSH24}
Akari Asai, Zeqiu Wu, Yizhong Wang, Avirup Sil, and Hannaneh Hajishirzi. 2024.
\newblock Self-rag: Learning to retrieve, generate, and critique through self-reflection.
\newblock In \emph{The Twelfth International Conference on Learning Representations, {ICLR} 2024, Vienna, Austria, May 7-11, 2024}. OpenReview.net.

\bibitem[{Brown et~al.(2024)Brown, Juravsky, Ehrlich, Clark, Le, R{\'{e}}, and Mirhoseini}]{abs-2407-21787}
Bradley C.~A. Brown, Jordan Juravsky, Ryan~Saul Ehrlich, Ronald Clark, Quoc~V. Le, Christopher R{\'{e}}, and Azalia Mirhoseini. 2024.
\newblock \href {https://doi.org/10.48550/ARXIV.2407.21787} {Large language monkeys: Scaling inference compute with repeated sampling}.
\newblock \emph{CoRR}, abs/2407.21787.

\bibitem[{Brown et~al.(2020)Brown, Mann, Ryder, Subbiah, Kaplan, Dhariwal, Neelakantan, Shyam, Sastry, Askell, Agarwal, Herbert{-}Voss, Krueger, Henighan, Child, Ramesh, Ziegler, Wu, Winter, Hesse, Chen, Sigler, Litwin, Gray, Chess, Clark, Berner, McCandlish, Radford, Sutskever, and Amodei}]{gpt3}
Tom~B. Brown, Benjamin Mann, Nick Ryder, Melanie Subbiah, Jared Kaplan, Prafulla Dhariwal, Arvind Neelakantan, Pranav Shyam, Girish Sastry, Amanda Askell, Sandhini Agarwal, Ariel Herbert{-}Voss, Gretchen Krueger, Tom Henighan, Rewon Child, Aditya Ramesh, Daniel~M. Ziegler, Jeffrey Wu, Clemens Winter, Christopher Hesse, Mark Chen, Eric Sigler, Mateusz Litwin, Scott Gray, Benjamin Chess, Jack Clark, Christopher Berner, Sam McCandlish, Alec Radford, Ilya Sutskever, and Dario Amodei. 2020.
\newblock Language models are few-shot learners.
\newblock In \emph{Advances in Neural Information Processing Systems 33: Annual Conference on Neural Information Processing Systems 2020, NeurIPS 2020, December 6-12, 2020, virtual}.

\bibitem[{Browne et~al.(2012)Browne, Powley, Whitehouse, Lucas, Cowling, Rohlfshagen, Tavener, Liebana, Samothrakis, and Colton}]{BrownePWLCRTPSC12}
Cameron Browne, Edward~Jack Powley, Daniel Whitehouse, Simon~M. Lucas, Peter~I. Cowling, Philipp Rohlfshagen, Stephen Tavener, Diego~Perez Liebana, Spyridon Samothrakis, and Simon Colton. 2012.
\newblock A survey of {Monte Carlo} tree search methods.
\newblock \emph{{IEEE} Trans. Comput. Intell. {AI} Games}, 4(1):1--43.

\bibitem[{Chen et~al.(2023)Chen, Shu, Shareghi, Collier, Narasimhan, and Yao}]{fireact}
Baian Chen, Chang Shu, Ehsan Shareghi, Nigel Collier, Karthik Narasimhan, and Shunyu Yao. 2023.
\newblock \href {https://doi.org/10.48550/ARXIV.2310.05915} {Fireact: Toward language agent fine-tuning}.
\newblock \emph{CoRR}, abs/2310.05915.

\bibitem[{Fan et~al.(2019)Fan, Jernite, Perez, Grangier, Weston, and Auli}]{FanJPGWA19}
Angela Fan, Yacine Jernite, Ethan Perez, David Grangier, Jason Weston, and Michael Auli. 2019.
\newblock {ELI5:} long form question answering.
\newblock In \emph{Proceedings of the 57th Conference of the Association for Computational Linguistics, {ACL} 2019, Florence, Italy, July 28- August 2, 2019, Volume 1: Long Papers}, pages 3558--3567. Association for Computational Linguistics.

\bibitem[{Fierro et~al.(2024)Fierro, Amplayo, Huot, Cao, Maynez, Narayan, and Lapata}]{FierroAHCMNL24}
Constanza Fierro, Reinald~Kim Amplayo, Fantine Huot, Nicola~De Cao, Joshua Maynez, Shashi Narayan, and Mirella Lapata. 2024.
\newblock Learning to plan and generate text with citations.
\newblock In \emph{Proceedings of the 62nd Annual Meeting of the Association for Computational Linguistics (Volume 1: Long Papers), {ACL} 2024, Bangkok, Thailand, August 11-16, 2024}, pages 11397--11417. Association for Computational Linguistics.

\bibitem[{Gao et~al.(2023{\natexlab{a}})Gao, Dai, Pasupat, Chen, Chaganty, Fan, Zhao, Lao, Lee, Juan, and Guu}]{GaoDPCCFZLLJG23}
Luyu Gao, Zhuyun Dai, Panupong Pasupat, Anthony Chen, Arun~Tejasvi Chaganty, Yicheng Fan, Vincent~Y. Zhao, Ni~Lao, Hongrae Lee, Da{-}Cheng Juan, and Kelvin Guu. 2023{\natexlab{a}}.
\newblock {RARR:} researching and revising what language models say, using language models.
\newblock In \emph{Proceedings of the 61st Annual Meeting of the Association for Computational Linguistics (Volume 1: Long Papers), {ACL} 2023, Toronto, Canada, July 9-14, 2023}, pages 16477--16508. Association for Computational Linguistics.

\bibitem[{Gao et~al.(2023{\natexlab{b}})Gao, Yen, Yu, and Chen}]{GaoYYC23}
Tianyu Gao, Howard Yen, Jiatong Yu, and Danqi Chen. 2023{\natexlab{b}}.
\newblock Enabling large language models to generate text with citations.
\newblock In \emph{Proceedings of the 2023 Conference on Empirical Methods in Natural Language Processing, {EMNLP} 2023, Singapore, December 6-10, 2023}, pages 6465--6488. Association for Computational Linguistics.

\bibitem[{Hart et~al.(1968)Hart, Nilsson, and Raphael}]{HartNR68}
Peter~E. Hart, Nils~J. Nilsson, and Bertram Raphael. 1968.
\newblock A formal basis for the heuristic determination of minimum cost paths.
\newblock \emph{{IEEE} Trans. Syst. Sci. Cybern.}, 4(2):100--107.

\bibitem[{Honovich et~al.(2022)Honovich, Aharoni, Herzig, Taitelbaum, Kukliansky, Cohen, Scialom, Szpektor, Hassidim, and Matias}]{HonovichAHTKCSS22}
Or~Honovich, Roee Aharoni, Jonathan Herzig, Hagai Taitelbaum, Doron Kukliansky, Vered Cohen, Thomas Scialom, Idan Szpektor, Avinatan Hassidim, and Yossi Matias. 2022.
\newblock {TRUE:} re-evaluating factual consistency evaluation.
\newblock In \emph{Proceedings of the 2022 Conference of the North American Chapter of the Association for Computational Linguistics: Human Language Technologies, {NAACL} 2022, Seattle, WA, United States, July 10-15, 2022}, pages 3905--3920. Association for Computational Linguistics.

\bibitem[{Hosseini et~al.(2024)Hosseini, Yuan, Malkin, Courville, Sordoni, and Agarwal}]{abs-2402-06457}
Arian Hosseini, Xingdi Yuan, Nikolay Malkin, Aaron~C. Courville, Alessandro Sordoni, and Rishabh Agarwal. 2024.
\newblock \href {https://doi.org/10.48550/ARXIV.2402.06457} {{V-STaR}: Training verifiers for self-taught reasoners}.
\newblock \emph{CoRR}, abs/2402.06457.

\bibitem[{Huang et~al.(2024)Huang, Wu, Hu, and Wang}]{HuangWHW24}
Chengyu Huang, Zeqiu Wu, Yushi Hu, and Wenya Wang. 2024.
\newblock Training language models to generate text with citations via fine-grained rewards.
\newblock In \emph{Proceedings of the 62nd Annual Meeting of the Association for Computational Linguistics (Volume 1: Long Papers), {ACL} 2024, Bangkok, Thailand, August 11-16, 2024}, pages 2926--2949. Association for Computational Linguistics.

\bibitem[{Huang et~al.(2023)Huang, Yu, Ma, Zhong, Feng, Wang, Chen, Peng, Feng, Qin, and Liu}]{abs-2311-05232}
Lei Huang, Weijiang Yu, Weitao Ma, Weihong Zhong, Zhangyin Feng, Haotian Wang, Qianglong Chen, Weihua Peng, Xiaocheng Feng, Bing Qin, and Ting Liu. 2023.
\newblock \href {https://doi.org/10.48550/ARXIV.2311.05232} {A survey on hallucination in large language models: Principles, taxonomy, challenges, and open questions}.
\newblock \emph{CoRR}, abs/2311.05232.

\bibitem[{Ji et~al.(2023)Ji, Lee, Frieske, Yu, Su, Xu, Ishii, Bang, Madotto, and Fung}]{JiLFYSXIBMF23}
Ziwei Ji, Nayeon Lee, Rita Frieske, Tiezheng Yu, Dan Su, Yan Xu, Etsuko Ishii, Yejin Bang, Andrea Madotto, and Pascale Fung. 2023.
\newblock Survey of hallucination in natural language generation.
\newblock \emph{{ACM} Comput. Surv.}, 55(12):248:1--248:38.

\bibitem[{Jiang et~al.(2023)Jiang, Xu, Gao, Sun, Liu, Dwivedi{-}Yu, Yang, Callan, and Neubig}]{JiangXGSLDYCN23}
Zhengbao Jiang, Frank~F. Xu, Luyu Gao, Zhiqing Sun, Qian Liu, Jane Dwivedi{-}Yu, Yiming Yang, Jamie Callan, and Graham Neubig. 2023.
\newblock Active retrieval augmented generation.
\newblock In \emph{Proceedings of the 2023 Conference on Empirical Methods in Natural Language Processing, {EMNLP} 2023, Singapore, December 6-10, 2023}, pages 7969--7992. Association for Computational Linguistics.

\bibitem[{Kahneman(2011)}]{kahneman2011thinking}
Daniel Kahneman. 2011.
\newblock Thinking, fast and slow.
\newblock \emph{Farrar, Straus and Giroux}.

\bibitem[{Kocsis and Szepesv{\'{a}}ri(2006)}]{uct}
Levente Kocsis and Csaba Szepesv{\'{a}}ri. 2006.
\newblock Bandit based {Monte-Carlo} planning.
\newblock In \emph{Machine Learning: {ECML} 2006, 17th European Conference on Machine Learning, Berlin, Germany, September 18-22, 2006, Proceedings}, volume 4212 of \emph{Lecture Notes in Computer Science}, pages 282--293. Springer.

\bibitem[{Lewis et~al.(2020)Lewis, Perez, Piktus, Petroni, Karpukhin, Goyal, K{\"{u}}ttler, Lewis, Yih, Rockt{\"{a}}schel, Riedel, and Kiela}]{LewisPPPKGKLYR020}
Patrick S.~H. Lewis, Ethan Perez, Aleksandra Piktus, Fabio Petroni, Vladimir Karpukhin, Naman Goyal, Heinrich K{\"{u}}ttler, Mike Lewis, Wen{-}tau Yih, Tim Rockt{\"{a}}schel, Sebastian Riedel, and Douwe Kiela. 2020.
\newblock Retrieval-augmented generation for knowledge-intensive {NLP} tasks.
\newblock In \emph{Advances in Neural Information Processing Systems 33: Annual Conference on Neural Information Processing Systems 2020, NeurIPS 2020, December 6-12, 2020, virtual}.

\bibitem[{Li et~al.(2024)Li, Sun, Hu, Liu, Hu, Liu, and Zhang}]{LiSHLH0Z24}
Dongfang Li, Zetian Sun, Baotian Hu, Zhenyu Liu, Xinshuo Hu, Xuebo Liu, and Min Zhang. 2024.
\newblock Improving attributed text generation of large language models via preference learning.
\newblock In \emph{Findings of the Association for Computational Linguistics, {ACL} 2024, Bangkok, Thailand and virtual meeting, August 11-16, 2024}, pages 5079--5101. Association for Computational Linguistics.

\bibitem[{Lightman et~al.(2024)Lightman, Kosaraju, Burda, Edwards, Baker, Lee, Leike, Schulman, Sutskever, and Cobbe}]{LightmanKBEBLLS24}
Hunter Lightman, Vineet Kosaraju, Yuri Burda, Harrison Edwards, Bowen Baker, Teddy Lee, Jan Leike, John Schulman, Ilya Sutskever, and Karl Cobbe. 2024.
\newblock Let's verify step by step.
\newblock In \emph{The Twelfth International Conference on Learning Representations, {ICLR} 2024, Vienna, Austria, May 7-11, 2024}. OpenReview.net.

\bibitem[{Liu et~al.(2023)Liu, Zhang, and Liang}]{LiuZL23}
Nelson~F. Liu, Tianyi Zhang, and Percy Liang. 2023.
\newblock Evaluating verifiability in generative search engines.
\newblock In \emph{Findings of the Association for Computational Linguistics: {EMNLP} 2023, Singapore, December 6-10, 2023}, pages 7001--7025. Association for Computational Linguistics.

\bibitem[{Narayan et~al.(2023)Narayan, Maynez, Amplayo, Ganchev, Louis, Huot, Sandholm, Das, and Lapata}]{NarayanMAGLH00L23}
Shashi Narayan, Joshua Maynez, Reinald~Kim Amplayo, Kuzman Ganchev, Annie Louis, Fantine Huot, Anders Sandholm, Dipanjan Das, and Mirella Lapata. 2023.
\newblock Conditional generation with a question-answering blueprint.
\newblock \emph{Trans. Assoc. Comput. Linguistics}, 11:974--996.

\bibitem[{Ni et~al.(2022)Ni, Qu, Lu, Dai, {\'{A}}brego, Ma, Zhao, Luan, Hall, Chang, and Yang}]{gtr}
Jianmo Ni, Chen Qu, Jing Lu, Zhuyun Dai, Gustavo~Hern{\'{a}}ndez {\'{A}}brego, Ji~Ma, Vincent~Y. Zhao, Yi~Luan, Keith~B. Hall, Ming{-}Wei Chang, and Yinfei Yang. 2022.
\newblock Large dual encoders are generalizable retrievers.
\newblock In \emph{Proceedings of the 2022 Conference on Empirical Methods in Natural Language Processing, {EMNLP} 2022, Abu Dhabi, United Arab Emirates, December 7-11, 2022}, pages 9844--9855. Association for Computational Linguistics.

\bibitem[{Piktus et~al.(2021)Piktus, Petroni, Karpukhin, Okhonko, Broscheit, Izacard, Lewis, Oguz, Grave, Yih, and Riedel}]{sphere}
Aleksandra Piktus, Fabio Petroni, Vladimir Karpukhin, Dmytro Okhonko, Samuel Broscheit, Gautier Izacard, Patrick S.~H. Lewis, Barlas Oguz, Edouard Grave, Wen{-}tau Yih, and Sebastian Riedel. 2021.
\newblock \href {http://arxiv.org/abs/2112.09924} {The web is your oyster - knowledge-intensive {NLP} against a very large web corpus}.
\newblock \emph{CoRR}, abs/2112.09924.

\bibitem[{Rafailov et~al.(2023)Rafailov, Sharma, Mitchell, Manning, Ermon, and Finn}]{dpo}
Rafael Rafailov, Archit Sharma, Eric Mitchell, Christopher~D. Manning, Stefano Ermon, and Chelsea Finn. 2023.
\newblock Direct preference optimization: Your language model is secretly a reward model.
\newblock In \emph{Advances in Neural Information Processing Systems 36: Annual Conference on Neural Information Processing Systems 2023, NeurIPS 2023, New Orleans, LA, USA, December 10 - 16, 2023}.

\bibitem[{Robertson et~al.(2009)Robertson, Zaragoza et~al.}]{bm25}
Stephen Robertson, Hugo Zaragoza, et~al. 2009.
\newblock The probabilistic relevance framework: {BM25} and beyond.
\newblock \emph{Foundations and Trends{\textregistered} in Information Retrieval}, 3(4):333--389.

\bibitem[{Shao et~al.(2023)Shao, Gong, Shen, Huang, Duan, and Chen}]{ShaoGSHDC23}
Zhihong Shao, Yeyun Gong, Yelong Shen, Minlie Huang, Nan Duan, and Weizhu Chen. 2023.
\newblock Enhancing retrieval-augmented large language models with iterative retrieval-generation synergy.
\newblock In \emph{Findings of the Association for Computational Linguistics: {EMNLP} 2023, Singapore, December 6-10, 2023}, pages 9248--9274. Association for Computational Linguistics.

\bibitem[{Shinn et~al.(2023)Shinn, Cassano, Gopinath, Narasimhan, and Yao}]{ShinnCGNY23}
Noah Shinn, Federico Cassano, Ashwin Gopinath, Karthik Narasimhan, and Shunyu Yao. 2023.
\newblock Reflexion: language agents with verbal reinforcement learning.
\newblock In \emph{Advances in Neural Information Processing Systems 36: Annual Conference on Neural Information Processing Systems 2023, NeurIPS 2023, New Orleans, LA, USA, December 10 - 16, 2023}.

\bibitem[{Silver et~al.(2016)Silver, Huang, Maddison, Guez, Sifre, van~den Driessche, Schrittwieser, Antonoglou, Panneershelvam, Lanctot, Dieleman, Grewe, Nham, Kalchbrenner, Sutskever, Lillicrap, Leach, Kavukcuoglu, Graepel, and Hassabis}]{alphago}
David Silver, Aja Huang, Chris~J. Maddison, Arthur Guez, Laurent Sifre, George van~den Driessche, Julian Schrittwieser, Ioannis Antonoglou, Vedavyas Panneershelvam, Marc Lanctot, Sander Dieleman, Dominik Grewe, John Nham, Nal Kalchbrenner, Ilya Sutskever, Timothy~P. Lillicrap, Madeleine Leach, Koray Kavukcuoglu, Thore Graepel, and Demis Hassabis. 2016.
\newblock Mastering the game of {Go} with deep neural networks and tree search.
\newblock \emph{Nat.}, 529(7587):484--489.

\bibitem[{Slobodkin et~al.(2024)Slobodkin, Hirsch, Cattan, Schuster, and Dagan}]{SlobodkinHCSD24}
Aviv Slobodkin, Eran Hirsch, Arie Cattan, Tal Schuster, and Ido Dagan. 2024.
\newblock Attribute first, then generate: Locally-attributable grounded text generation.
\newblock In \emph{Proceedings of the 62nd Annual Meeting of the Association for Computational Linguistics (Volume 1: Long Papers), {ACL} 2024, Bangkok, Thailand, August 11-16, 2024}, pages 3309--3344. Association for Computational Linguistics.

\bibitem[{Snell et~al.(2024)Snell, Lee, Xu, and Kumar}]{abs-2408-03314}
Charlie Snell, Jaehoon Lee, Kelvin Xu, and Aviral Kumar. 2024.
\newblock Scaling {LLM} test-time compute optimally can be more effective than scaling model parameters.
\newblock \emph{CoRR}, abs/2408.03314.

\bibitem[{Stelmakh et~al.(2022)Stelmakh, Luan, Dhingra, and Chang}]{StelmakhLDC22}
Ivan Stelmakh, Yi~Luan, Bhuwan Dhingra, and Ming{-}Wei Chang. 2022.
\newblock {ASQA:} factoid questions meet long-form answers.
\newblock In \emph{Proceedings of the 2022 Conference on Empirical Methods in Natural Language Processing, {EMNLP} 2022, Abu Dhabi, United Arab Emirates, December 7-11, 2022}, pages 8273--8288. Association for Computational Linguistics.

\bibitem[{Sun et~al.(2024)Sun, Cai, Wang, Hou, Wei, Wang, Zhang, and Yin}]{SunCWHWWZY24}
Hao Sun, Hengyi Cai, Bo~Wang, Yingyan Hou, Xiaochi Wei, Shuaiqiang Wang, Yan Zhang, and Dawei Yin. 2024.
\newblock Towards verifiable text generation with evolving memory and self-reflection.
\newblock In \emph{Proceedings of the 2024 Conference on Empirical Methods in Natural Language Processing, {EMNLP} 2024, Miami, FL, USA, November 12-16, 2024}, pages 8211--8227. Association for Computational Linguistics.

\bibitem[{Sutton(2019)}]{sutton2019bitter}
Richard Sutton. 2019.
\newblock The bitter lesson.
\newblock \emph{Incomplete Ideas (blog)}, 13(1):38.

\bibitem[{Touvron et~al.(2023)Touvron, Martin, Stone, Albert, Almahairi, Babaei, Bashlykov, Batra, Bhargava, Bhosale, Bikel, Blecher, Canton{-}Ferrer, Chen, Cucurull, Esiobu, Fernandes, Fu, Fu, Fuller, Gao, Goswami, Goyal, Hartshorn, Hosseini, Hou, Inan, Kardas, Kerkez, Khabsa, Kloumann, Korenev, Koura, Lachaux, Lavril, Lee, Liskovich, Lu, Mao, Martinet, Mihaylov, Mishra, Molybog, Nie, Poulton, Reizenstein, Rungta, Saladi, Schelten, Silva, Smith, Subramanian, Tan, Tang, Taylor, Williams, Kuan, Xu, Yan, Zarov, Zhang, Fan, Kambadur, Narang, Rodriguez, Stojnic, Edunov, and Scialom}]{llama2}
Hugo Touvron, Louis Martin, Kevin Stone, Peter Albert, Amjad Almahairi, Yasmine Babaei, Nikolay Bashlykov, Soumya Batra, Prajjwal Bhargava, Shruti Bhosale, Dan Bikel, Lukas Blecher, Cristian Canton{-}Ferrer, Moya Chen, Guillem Cucurull, David Esiobu, Jude Fernandes, Jeremy Fu, Wenyin Fu, Brian Fuller, Cynthia Gao, Vedanuj Goswami, Naman Goyal, Anthony Hartshorn, Saghar Hosseini, Rui Hou, Hakan Inan, Marcin Kardas, Viktor Kerkez, Madian Khabsa, Isabel Kloumann, Artem Korenev, Punit~Singh Koura, Marie{-}Anne Lachaux, Thibaut Lavril, Jenya Lee, Diana Liskovich, Yinghai Lu, Yuning Mao, Xavier Martinet, Todor Mihaylov, Pushkar Mishra, Igor Molybog, Yixin Nie, Andrew Poulton, Jeremy Reizenstein, Rashi Rungta, Kalyan Saladi, Alan Schelten, Ruan Silva, Eric~Michael Smith, Ranjan Subramanian, Xiaoqing~Ellen Tan, Binh Tang, Ross Taylor, Adina Williams, Jian~Xiang Kuan, Puxin Xu, Zheng Yan, Iliyan Zarov, Yuchen Zhang, Angela Fan, Melanie Kambadur, Sharan Narang, Aur{\'{e}}lien Rodriguez, Robert Stojnic, Sergey Edunov,
  and Thomas Scialom. 2023.
\newblock Llama 2: Open foundation and fine-tuned chat models.
\newblock \emph{CoRR}, abs/2307.09288.

\bibitem[{Wang et~al.(2024)Wang, Deng, Lv, Liang, He, Yan, and An}]{abs-2406-14283}
Chaojie Wang, Yanchen Deng, Zhiyi Lv, Zeng Liang, Jujie He, Shuicheng Yan, and Bo~An. 2024.
\newblock \href {https://doi.org/10.48550/ARXIV.2406.14283} {Q*: Improving multi-step reasoning for {LLMs} with deliberative planning}.
\newblock \emph{CoRR}, abs/2406.14283.

\bibitem[{Yao et~al.(2022)Yao, Zhao, Yu, Du, Shafran, Narasimhan, and Cao}]{Yao-arxiv-2022-ReAct}
Shunyu Yao, Jeffrey Zhao, Dian Yu, Nan Du, Izhak Shafran, Karthik Narasimhan, and Yuan Cao. 2022.
\newblock React: Synergizing reasoning and acting in language models.
\newblock \emph{arXiv preprint arXiv:2210.03629}.

\bibitem[{Ye and Ng(2024)}]{ye-ng-2024-preference}
Hai Ye and Hwee~Tou Ng. 2024.
\newblock \href {https://doi.org/10.18653/v1/2024.emnlp-main.1206} {Preference-guided reflective sampling for aligning language models}.
\newblock In \emph{Proceedings of the 2024 Conference on Empirical Methods in Natural Language Processing}, pages 21646--21668, Miami, Florida, USA. Association for Computational Linguistics.

\bibitem[{Ye et~al.(2021)Ye, Liu, Kurutach, Abbeel, and Gao}]{YeLKAG21}
Weirui Ye, Shaohuai Liu, Thanard Kurutach, Pieter Abbeel, and Yang Gao. 2021.
\newblock Mastering {Atari} games with limited data.
\newblock In \emph{NeurIPS}, pages 25476--25488.

\bibitem[{Yu et~al.(2024)Yu, Peng, Vajipey, Cheng, Galley, Gao, and Yu}]{yu2024improving}
Xiao Yu, Baolin Peng, Vineeth Vajipey, Hao Cheng, Michel Galley, Jianfeng Gao, and Zhou Yu. 2024.
\newblock Improving autonomous {AI} agents with reflective tree search and self-learning.
\newblock \emph{arXiv preprint arXiv:2410.02052}.

\bibitem[{Zhang et~al.(2024)Zhang, Wu, Lei, Che, Li, Xie, Huang, Zhang, Pavone, Li et~al.}]{zhang2024llama}
Di~Zhang, Jianbo Wu, Jingdi Lei, Tong Che, Jiatong Li, Tong Xie, Xiaoshui Huang, Shufei Zhang, Marco Pavone, Yuqiang Li, et~al. 2024.
\newblock {LLaMA-Berry}: Pairwise optimization for {O1-like} {Olympiad-Level} mathematical reasoning.
\newblock \emph{arXiv preprint arXiv:2410.02884}.

\bibitem[{Zhang et~al.(2023)Zhang, Li, Cui, Cai, Liu, Fu, Huang, Zhao, Zhang, Chen, Wang, Luu, Bi, Shi, and Shi}]{abs-2309-01219}
Yue Zhang, Yafu Li, Leyang Cui, Deng Cai, Lemao Liu, Tingchen Fu, Xinting Huang, Enbo Zhao, Yu~Zhang, Yulong Chen, Longyue Wang, Anh~Tuan Luu, Wei Bi, Freda Shi, and Shuming Shi. 2023.
\newblock \href {https://doi.org/10.48550/ARXIV.2309.01219} {Siren's song in the {AI} ocean: {A} survey on hallucination in large language models}.
\newblock \emph{CoRR}, abs/2309.01219.

\bibitem[{Zhao et~al.(2023)Zhao, Zhou, Li, Tang, Wang, Hou, Min, Zhang, Zhang, Dong, Du, Yang, Chen, Chen, Jiang, Ren, Li, Tang, Liu, Liu, Nie, and Wen}]{llm_survey}
Wayne~Xin Zhao, Kun Zhou, Junyi Li, Tianyi Tang, Xiaolei Wang, Yupeng Hou, Yingqian Min, Beichen Zhang, Junjie Zhang, Zican Dong, Yifan Du, Chen Yang, Yushuo Chen, Zhipeng Chen, Jinhao Jiang, Ruiyang Ren, Yifan Li, Xinyu Tang, Zikang Liu, Peiyu Liu, Jian{-}Yun Nie, and Ji{-}Rong Wen. 2023.
\newblock A survey of large language models.
\newblock \emph{CoRR}, abs/2303.18223.

\bibitem[{Zhou et~al.(2024)Zhou, Yan, Shlapentokh{-}Rothman, Wang, and Wang}]{ZhouYSWW24}
Andy Zhou, Kai Yan, Michal Shlapentokh{-}Rothman, Haohan Wang, and Yu{-}Xiong Wang. 2024.
\newblock Language agent tree search unifies reasoning, acting, and planning in language models.
\newblock In \emph{Forty-first International Conference on Machine Learning, {ICML} 2024, Vienna, Austria, July 21-27, 2024}. OpenReview.net.

\end{thebibliography}
\bibliographystyle{acl_natbib}

\newpage
\clearpage
\appendix

\section*{Appendix}
\label{appendix}

\section{Datasets}
\label{sec:app-dataset}

We evaluate our approach on the ALCE benchmark~\cite{GaoYYC23} consisting of three datasets. Specifically, the ASQA dataset~\citep{StelmakhLDC22} contains 948 questions, where the answers can be found from Wikipedia; the QAMPARI dataset~\citep{qampari} contains 1,000 questions based on Wikipedia; and the ELI5 dataset~\citep{FanJPGWA19} includes 1,000 questions, where the answers can be found from Sphere~\citep{sphere}. The details of these three datasets are given in Table~\ref{tab:dataset}. {Following the setting of the ALCE benchmark~\cite{GaoYYC23}, we divide the retrieval corpus into 100-word passages to enable a fair and consistent comparison to baselines. Besides, 100-word chunks provide more precise evidence for the generated content, make it easier for humans to verify, and do not introduce too much irrelevant information.}

\begin{table}[H]
    \centering\small
    \begin{tabular}{ccc}
        \toprule
        \textbf{Dataset} & \textbf{Corpus (\#Passages)} & \textbf{Question Type} \\
        \midrule
        ASQA & Wikipedia (21M) & Factoid \\
        QAMPARI & Wikipedia (21M) & Factoid (list) \\
        ELI5 & Sphere (899M) & Why/How/What \\
         \bottomrule
    \end{tabular}
    \caption{Details of the three evaluation datasets.}
    \label{tab:dataset}
\end{table}

\begin{figure*}[t]
    \centering
    \small
    \includegraphics[width=\textwidth]{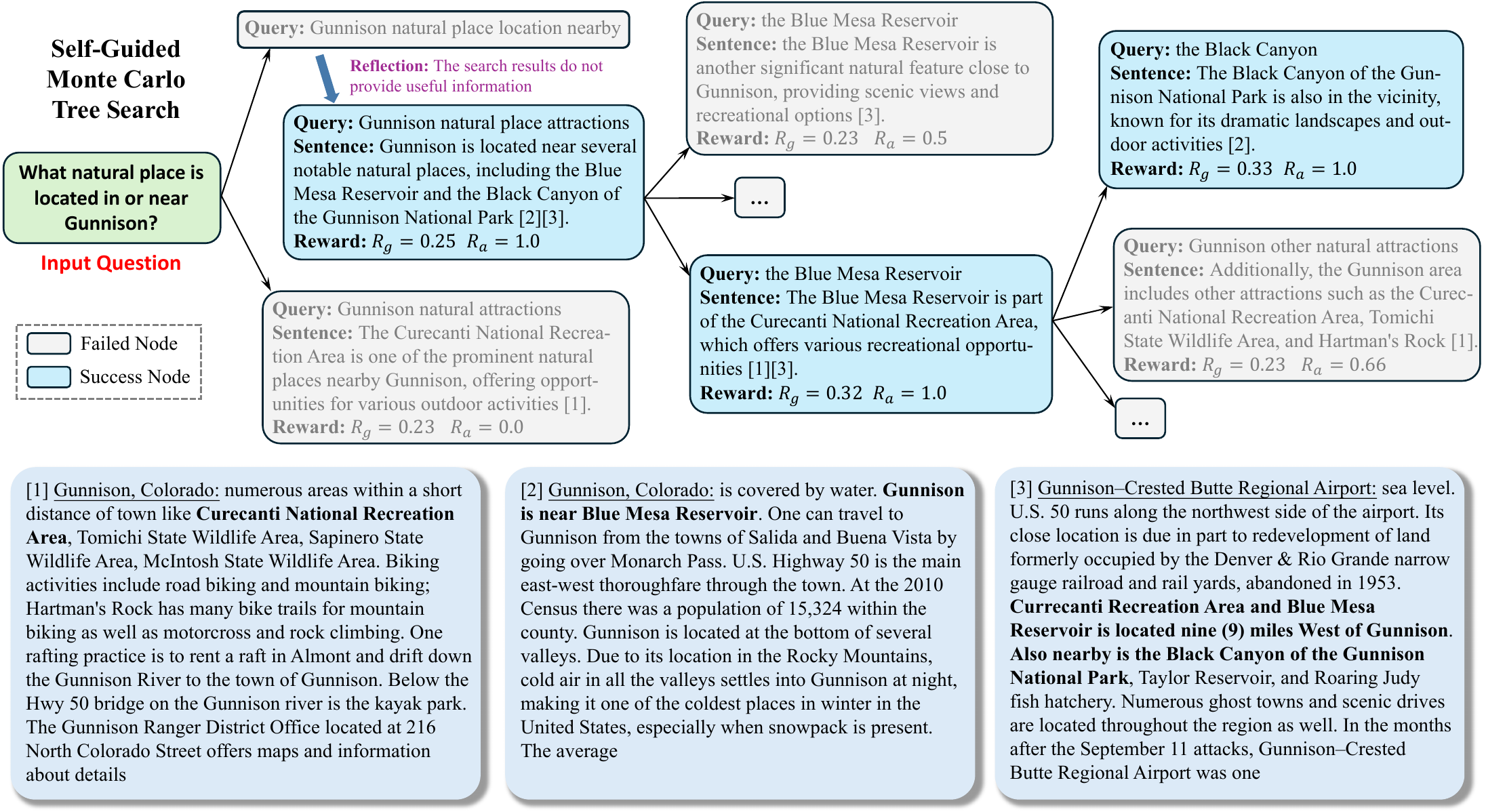}
    \caption{A qualitative example from ASQA showing the attributed generation process of Think\&Cite.
    } 
    \label{fig:case}
    \vspace{-0.3cm}
\end{figure*}

\begin{table}[t]
\setlength\tabcolsep{4pt}
\centering
\small
\begin{tabular}{l c c c c c}
    \toprule
    \textbf{Method} & \textbf{\makecell{NI}} & \textbf{\makecell{TT}} & \textbf{\makecell{NT}} & \textbf{Recall} & \textbf{Precision} \\
    
    \midrule
    FG-Reward  & - & 60 & 1229 & 77.8 & 76.3  \\
    VTG & - & 76 & 1057 & 86.7 & 80.0 \\
    \midrule
    \multirow{4}{*}{Think\&Cite~(Ours)} & 10 & 67 & 1149 & 83.5 & 81.6 \\
    & 20 & 78 & 1238 & 85.7 & 81.8 \\
    & 30 & 96 & 1547 & 89.5 & 87.1 \\
    & 40 & 103 & 1720 & 87.0 & 85.4 \\
    \bottomrule
\end{tabular}
\caption{Comparison of time costs per question and the citation recall and precision. Here, \textbf{NI} represents the maximum number of MCTS iterations per question, \textbf{TT} and \textbf{NT} represent the total inference time (second) and number of generated tokens per question, respectively.}
\label{tab:time}
\end{table}

\section{Analysis of Computational Expenses}

{We conduct further analysis of computational costs using our model (incorporating GPT-4o) concerning different numbers of MCTS iterations on the ASQA dataset. First, we observe that using our model, the average time taken and the average number of tokens increase when the number of MCTS iterations increases. Citation performance also improves (except when number of iterations = 40). These results are in line with recent results reported in the literature on test-time scaling. Second, because FG-Reward adopts rejection sampling (which involves sampling multiple times) to generate text and VTG uses the whole corpus to verify citation correctness, their total time costs are also high to some extent.}

\section{Prompts}
\label{sec:app-prompt}

We instruct the LLM to perform self-guided tree search for attributed text generation in a few-shot manner. The in-context learning prompts for the ASQA, QAMPARI, and ELI5 datasets are presented in Table~\ref{tab:asqa-instruction}, Table~\ref{tab:qampari-instruction}, and Table~\ref{tab:eli5-instruction} respectively. In these prompts, we first define four operations for the LLM in the iterative think-verbalize-cite paradigm. Then, we construct four in-context examples, which are selected to be consistent with the ALCE benchmark~\citep{GaoYYC23}.

\begin{table*}[t]
	\small
	\centering
	\begin{tabular}{p{0.97\textwidth}}
		\toprule
            \rowcolor{tBlue}{\fontfamily{ppl}\selectfont Your objective is to write an accurate, engaging, and concise answer for the given question using only the provided search results and cite them properly. Use an unbiased and journalistic tone. Always cite for any factual claim. You can use four operations:} \\
            \rowcolor{tBlue}
            {\fontfamily{ppl}\selectfont (1) """Search: key words""" which requests some key words to retrieve the most relevant documents based on the search and generation history.} \\
            \rowcolor{tBlue}
            {\fontfamily{ppl}\selectfont (2) """Reflexion: thoughts""" which reflects on the search results if not obtaining valid documents for answering the question.} \\
            \rowcolor{tBlue}
            {\fontfamily{ppl}\selectfont (3) """Output: sentences""" which outputs a long-form answer based on retrieved documents. In the answer, cite properly by using [1][2][3]. Cite at least one document and at most three documents in each sentence. If multiple documents support the sentence, only cite a minimum sufficient subset of the documents.} \\
            \rowcolor{tBlue}
            {\fontfamily{ppl}\selectfont (4) """End""" which ends the task of question answering.} \\
            \rowcolor{tBlue}
            {\fontfamily{ppl}\selectfont Here are some example:} \\
            \specialrule{0em}{1pt}{1pt}
            \rowcolor{tGreen}{\fontfamily{ppl}\selectfont \textbf{Question:} Who set the record for longest field goal?} \\
            \rowcolor{tGreen}{\fontfamily{ppl}\selectfont \textbf{Search:} record for longest field goal} \\
            \rowcolor{tGreen}{\fontfamily{ppl}\selectfont Document [1] (Title: David Akers) record for the longest field goal by kicking a 63-yard field goal off the crossbar ...} \\
            \rowcolor{tGreen}{\fontfamily{ppl}\selectfont Document [2] (Title: Dirk Borgognone) Dirk Borgognone Dirk Ronald Borgognone (born January 9, 1968) is a former National ...} \\
            \rowcolor{tGreen}{\fontfamily{ppl}\selectfont Document [3] (Title: Russell Erxleben) against the University of Colorado. Despite early troubles in the game, including ...} \\
            \rowcolor{tGreen}{\fontfamily{ppl}\selectfont \textbf{Reflexion:} The search results do not provide valid information to answer the question. It might be better to search the longest field goal in NFL.} \\
            \rowcolor{tGreen}{\fontfamily{ppl}\selectfont \textbf{Search:} record for longest field goal NFL} \\
            \rowcolor{tGreen}{\fontfamily{ppl}\selectfont Document [4] (Title: Field goal) toward its own end. The longest field goal kick in NFL history is 64 yards, a record set ...} \\
            \rowcolor{tGreen}{\fontfamily{ppl}\selectfont Document [5] (Title: Field goal) both end zones) is only 66 yards. Scaccia, while playing indoor football, attempted a 64-yard ...} \\
            \rowcolor{tGreen}{\fontfamily{ppl}\selectfont Document [6] (Title: Field goal) this accomplishment is not the official record. All of the above kicks were successful ...} \\
            \rowcolor{tGreen}{\fontfamily{ppl}\selectfont 
            \textbf{Output:} The record for the longest field goal in an NFL game was set by Matt Prater at 64 yards [4].} \\
            \rowcolor{tGreen}{\fontfamily{ppl}\selectfont \textbf{Search:} record for longest field goal at any level college} \\
            \rowcolor{tGreen}{\fontfamily{ppl}\selectfont Document [7] (Title: Field goal) toward its own end. The longest field goal kick in NFL history is 64 yards, a record set by ...} \\
            \rowcolor{tGreen}{\fontfamily{ppl}\selectfont Document [8] (Title: Field goal range) 35 and 40 yard lines (closer in a crosswind) often will go for the more risky ...} \\
            \rowcolor{tGreen}{\fontfamily{ppl}\selectfont Document [9] (Title: Field goal range) NFL and have been banned from NCAA since 1989) is 68 yards held by Fabrizio Scaccia ...} \\
            \rowcolor{tGreen}{\fontfamily{ppl}\selectfont 
            \textbf{Output:} But the record for the longest field goal at any level was 69 yards, kicked by collegiate kicker Ove Johansson in a 1976 Abilene Christian University football game against East Texas State University [8].} \\
            \rowcolor{tGreen}{\fontfamily{ppl}\selectfont 
            \textbf{End}} \\
            \specialrule{0em}{1pt}{1pt}
            \rowcolor{tGreen}{\fontfamily{ppl}\selectfont <Remaining Three Demonstrations>} \\
            \specialrule{0em}{1pt}{1pt}
            {\fontfamily{ppl}\selectfont Now, following the example's output format to solve this new question. } \\
            \\
            {\fontfamily{ppl}\selectfont \textbf{Question:} <insert the input question>} \\
            \bottomrule
	\end{tabular}
	\caption{Prompt of attributed text generation in ASQA. The \colorbox{tBlue}{blue} text denotes the intention description, and the \colorbox{tGreen}{green} text denotes the generation demonstration.} 
	\label{tab:asqa-instruction}
\end{table*}

\begin{table*}[t]
	\small
	\centering
	\begin{tabular}{p{0.97\textwidth}}
		\toprule
            \rowcolor{tBlue}{\fontfamily{ppl}\selectfont Your objective is to write an accurate, engaging, and concise answer for the given question using only the provided search results and cite them properly. Use an unbiased and journalistic tone. Always cite for any factual claim. You can use four operations:} \\
            \rowcolor{tBlue}
            {\fontfamily{ppl}\selectfont (1) """Search: key words""" which requests some key words to retrieve the most relevant documents based on the search and generation history.} \\
            \rowcolor{tBlue}
            {\fontfamily{ppl}\selectfont (2) """Reflexion: thoughts""" which reflects on the search results if not obtaining valid documents for answering the question.} \\
            \rowcolor{tBlue}
            {\fontfamily{ppl}\selectfont (3) """Output: sentences""" which outputs a long-form answer based on retrieved documents. In the answer, cite properly by using [1][2][3]. Cite at least one document and at most three documents in each sentence. If multiple documents support the sentence, only cite a minimum sufficient subset of the documents.} \\
            \rowcolor{tBlue}
            {\fontfamily{ppl}\selectfont (4) """End""" which ends the task of question answering.} \\
            \rowcolor{tBlue}
            {\fontfamily{ppl}\selectfont Here are some example:} \\
            \specialrule{0em}{1pt}{1pt}
            \rowcolor{tGreen}{\fontfamily{ppl}\selectfont \textbf{Question:} Which books were written by Nevil Shute?} \\
            \rowcolor{tGreen}{\fontfamily{ppl}\selectfont \textbf{Search:} Nevil Shute books theme social barriers} \\
            \rowcolor{tGreen}{\fontfamily{ppl}\selectfont Document [1] (Title: Nevil Shute) early stages. My congratulations." His celebrity as a writer caused the Ministry ...} \\
            \rowcolor{tGreen}{\fontfamily{ppl}\selectfont Document [2] (Title: Nevil Shute) theme is the bridging of social barriers such as class ("Lonely Road" and "Landfall"), race ...} \\
            \rowcolor{tGreen}{\fontfamily{ppl}\selectfont Document [3] (Title: Nevil Shute) construction company, Airspeed Ltd. His popularity grew slowly with each novel, but he became much more famous ...} \\
            \rowcolor{tGreen}{\fontfamily{ppl}\selectfont \textbf{Output:} Beyond the Black Stump [2], Lonely Road [2], The Chequer Board [2], In the Wet [2], Trustee from the Toolroom [2], Round the Bend [2].} \\
            \rowcolor{tGreen}{\fontfamily{ppl}\selectfont \textbf{Search:} Nevil Shute books simple readable style} \\
            \rowcolor{tGreen}{\fontfamily{ppl}\selectfont Document [4] (Nevil Shute) construction company, Airspeed Ltd. His popularity grew slowly with each novel, but he became much more famous ...} \\
            \rowcolor{tGreen}{\fontfamily{ppl}\selectfont Document [5] (The Chequer Board) the Burmese people", both of which are central to the book's story. Shute was concerned that sales of the book ...} \\
            \rowcolor{tGreen}{\fontfamily{ppl}\selectfont Document [6] (In the Wet) had used the idea of multiple votes for merit in his short story "The Curious Republic of Gondour". ...} \\
            \rowcolor{tGreen}{\fontfamily{ppl}\selectfont 
            \textbf{Reflexion:} The search results do not provide any useful information to answer the question. It might be better to search Nevil Shute books in 1950s.} \\
            \rowcolor{tGreen}{\fontfamily{ppl}\selectfont \textbf{Search:} Nevil Shute books 1950s} \\
            \rowcolor{tGreen}{\fontfamily{ppl}\selectfont Document [7] (Nevil Shute) early stages. My congratulations." His celebrity as a writer caused the Ministry of Information to send him to the ...} \\
            \rowcolor{tGreen}{\fontfamily{ppl}\selectfont Document [8] (Nevil Shute) theme is the bridging of social barriers such as class ("Lonely Road" and "Landfall"), race ("The Chequer Board") ...} \\
            \rowcolor{tGreen}{\fontfamily{ppl}\selectfont Document [9] (Nevil Shute) construction company, Airspeed Ltd. His popularity grew slowly with each novel, but he became much more famous ...} \\
            \rowcolor{tGreen}{\fontfamily{ppl}\selectfont 
            \textbf{Output:} Marazan [7], Stephen Morris [7].} \\
            \rowcolor{tGreen}{\fontfamily{ppl}\selectfont 
            \textbf{End}} \\
            \specialrule{0em}{1pt}{1pt}
            \rowcolor{tGreen}{\fontfamily{ppl}\selectfont <Remaining Three Demonstrations>} \\
            \specialrule{0em}{1pt}{1pt}
            {\fontfamily{ppl}\selectfont Now, following the example's output format to solve this new question. } \\
            \\
            {\fontfamily{ppl}\selectfont \textbf{Question:} <insert the input question>} \\
            \bottomrule
	\end{tabular}
	\caption{Prompt of attributed text generation in QAMPARI. The \colorbox{tBlue}{blue} text denotes the intention description, and the \colorbox{tGreen}{green} text denotes the generation demonstration.} 
	\label{tab:qampari-instruction}
\end{table*}

\begin{table*}[t]
	\small
	\centering
	\begin{tabular}{p{0.97\textwidth}}
		\toprule
            \rowcolor{tBlue}{\fontfamily{ppl}\selectfont Your objective is to write an accurate, engaging, and concise answer for the given question using only the provided search results and cite them properly. Use an unbiased and journalistic tone. Always cite for any factual claim. You can use four operations:} \\
            \rowcolor{tBlue}
            {\fontfamily{ppl}\selectfont (1) """Search: key words""" which requests some key words to retrieve the most relevant documents based on the search and generation history.} \\
            \rowcolor{tBlue}
            {\fontfamily{ppl}\selectfont (2) """Reflexion: thoughts""" which reflects on the search results if not obtaining valid documents for answering the question.} \\
            \rowcolor{tBlue}
            {\fontfamily{ppl}\selectfont (3) """Output: sentences""" which outputs a long-form answer based on retrieved documents. In the answer, cite properly by using [1][2][3]. Cite at least one document and at most three documents in each sentence. If multiple documents support the sentence, only cite a minimum sufficient subset of the documents.} \\
            \rowcolor{tBlue}
            {\fontfamily{ppl}\selectfont (4) """End""" which ends the task of question answering.} \\
            \rowcolor{tBlue}
            {\fontfamily{ppl}\selectfont Here are some example:} \\
            \specialrule{0em}{1pt}{1pt}
            \rowcolor{tGreen}{\fontfamily{ppl}\selectfont \textbf{Question:} What causes Bi-polar disorder?} \\
            \rowcolor{tGreen}{\fontfamily{ppl}\selectfont \textbf{Search:} What is Bi-polar disorder?} \\
            \rowcolor{tGreen}{\fontfamily{ppl}\selectfont Document [1] (Title: Bipolar II disorder) setting boundaries for one's perceptions and behaviors. Studies indicate that the following ...} \\
            \rowcolor{tGreen}{\fontfamily{ppl}\selectfont Document [2] (Title: Biology of bipolar disorder) located in these regions have been hypothesized as a mechanism for bipolar ...} \\
            \rowcolor{tGreen}{\fontfamily{ppl}\selectfont Document [3] (Title: Biology of bipolar disorder) regardless of polarity. Inflammation has been consistently reported in bipolar ...} \\
            \rowcolor{tGreen}{\fontfamily{ppl}\selectfont \textbf{Reflexion:} The search results do not provide valid information to answer the question. It might be better to search the symptoms of Bi-polar disorder.} \\
            \rowcolor{tGreen}{\fontfamily{ppl}\selectfont \textbf{Search:} What are the symptoms of Bi-polar disorder and how long do they last?} \\
            \rowcolor{tGreen}{\fontfamily{ppl}\selectfont Document [4] (Title: Bi-polar disorder | definition of Bi-polar disorder by Medical dictionary) bi-polar disorder | definition of bi-polar ...} \\
            \rowcolor{tGreen}{\fontfamily{ppl}\selectfont Document [5] (Title: Mania and Bi-Polar) can go from depressed to \u201csuper happy\u201d all in one day, or even in a few days, does ...} \\
            \rowcolor{tGreen}{\fontfamily{ppl}\selectfont Document [6] (Title: For Individuals — Adam Schwartz) For Individuals \u2014 Adam Schwartz The information is extensive and covers a ...} \\
            \rowcolor{tGreen}{\fontfamily{ppl}\selectfont 
            \textbf{Output:} Bipolar disorder is an emotional disorder that causes extreme mood swings between excitement and depression [4]. The spectrum of mood swing may span from days to months [5].} \\
            \rowcolor{tGreen}{\fontfamily{ppl}\selectfont \textbf{Search:} What could cuase Bi-polar disorder?} \\
            \rowcolor{tGreen}{\fontfamily{ppl}\selectfont Document [7] (Title: Bi-Polar disorder) Bi-Polar disorder Bi-polar is generally a cyclic disease where individuals display depressive ...} \\
            \rowcolor{tGreen}{\fontfamily{ppl}\selectfont Document [8] (Title: Depression Bi-polar Disorder Symptoms 2019 | Win Over Depression) Depression Bi-polar Disorder Symptoms 2019 ...} \\
            \rowcolor{tGreen}{\fontfamily{ppl}\selectfont Document [9] (Title: Mental disorder) may include parenting factors including parental rejection, lack of parental warmth ...} \\
            \rowcolor{tGreen}{\fontfamily{ppl}\selectfont 
            \textbf{Output:} We are still not certain of the exact factors that cause such disorder, but genetics is considered a major factor [7].} \\
            \rowcolor{tGreen}{\fontfamily{ppl}\selectfont 
            \textbf{End}} \\
            \specialrule{0em}{1pt}{1pt}
            \rowcolor{tGreen}{\fontfamily{ppl}\selectfont <Remaining Three Demonstrations>} \\
            \specialrule{0em}{1pt}{1pt}
            {\fontfamily{ppl}\selectfont Now, following the example's output format to solve this new question. } \\
            \\
            {\fontfamily{ppl}\selectfont \textbf{Question:} <insert the input question>} \\
            \bottomrule
	\end{tabular}
	\caption{Prompt of attributed text generation in ELI5. The \colorbox{tBlue}{blue} text denotes the intention description, and the \colorbox{tGreen}{green} text denotes the generation demonstration.} 
	\label{tab:eli5-instruction}
\end{table*}

\section{Case Study}
\label{sec:app-case}

We present an example from ASQA in Figure~\ref{fig:case}.

\clearpage

\end{document}